\pdfoutput=1
\documentclass[11pt]{article}

\usepackage[]{authblk}
\usepackage{mdframed}
\usepackage{setspace}
\usepackage{multirow}
\usepackage{amsmath} 
\usepackage{multirow}
\usepackage{booktabs}
\usepackage{cjhebrew}
\usepackage{hhline}
\usepackage{CJKutf8}

\usepackage{xspace}
\include{amsmath}

\usepackage{subfig}

\usepackage{graphicx}
\usepackage[T1]{fontenc}
\usepackage{inputenc}
\usepackage{microtype}

\usepackage{inconsolata}
\usepackage{csquotes}

\usepackage{todonotes}

\makeatletter
\renewcommand\AB@affilsepx{, \protect\Affilfont}
\makeatother

\usepackage[]{ACL2023}
\let\oldfootnotetext\footnotetext
\renewcommand{\footnotetext}[1]{%
  \begingroup%
  \renewcommand{\thefootnote}{\ensuremath{*}}%
  \oldfootnotetext{#1}%
  \endgroup%
}

\usepackage{times}

\usepackage{booktabs}
\usepackage{siunitx}
\usepackage{latexsym}
\usepackage{graphicx}
\usepackage{tabu}
\usepackage{multirow}
\usepackage{makecell} 

\usepackage[T1]{fontenc}
\usepackage{microtype}

\usepackage{inconsolata}
\usepackage{csquotes}

\usepackage{todonotes}

\newcommand{\pz}{\hphantom{0}}
\newcommand{\pzz}{\hphantom{00}}

\author[a]{Itai Mondshine}
\author[a]{Tzuf Paz-Argaman}
\author[a]{Reut Tsarfaty}

\affil[a]{Bar-Ilan University, Israel}
\affil[ ]{\authorcr \tt \{mondshi1, tzuf.paz-argaman, reut.tsarfaty\}@biu.ac.il}

\title{Beyond English: The Impact of Prompt Translation Strategies \\across Languages and Tasks in Multilingual LLMs }

\begin{document}
\maketitle
\begin{abstract}
Despite advances in the multilingual capabilities of Large Language Models (LLMs) across diverse 
%Natural Language Processing (NLP)
tasks, English remains the dominant language for LLM research and development. This
has led to the widespread practice of \emph{pre-translation}, i.e., translating non-English task prompts into English before inference. \emph{Selective pre-translation}, a more surgical approach, focuses on translating specific prompt components. However, its current use is sporadic and lacks a systematic research foundation. Consequently, the optimal \emph{selective pre-translation} strategy for various multilingual settings and tasks remains unclear. In this work, we aim to uncover the optimal setup for \emph{selective pre-translation} by systematically assessing its use. Specifically, we view the prompt as a modular entity, composed of four functional parts: instruction, context, examples, and output, either of which could be translated or not.  We evaluate pre-translation strategies across 35 languages covering both low and high-resource languages, on various tasks including Question Answering (QA), Natural Language Inference (NLI), Named Entity Recognition (NER), and Abstractive Summarization. Our experiments show the impact of factors as similarity to English, translation quality, and the size of pre-trained data, on the model performance. We suggest practical guidelines for choosing  optimal strategies in various multilingual settings.\footnote{
We launched a user-friendly HuggingFace Space for generation and use of selective pre-translation prompts 
%and facilitating implementation: 
\url{https://huggingface.co/spaces/naacl-anonymous/selective_pre_translation}. Appendix \ref{appendix:hf_space} provides further details and illustrations.}

\end{abstract}

\begin{figure*}[th]
    \centering
    \scalebox{0.85}{
    \includegraphics[width=\textwidth]{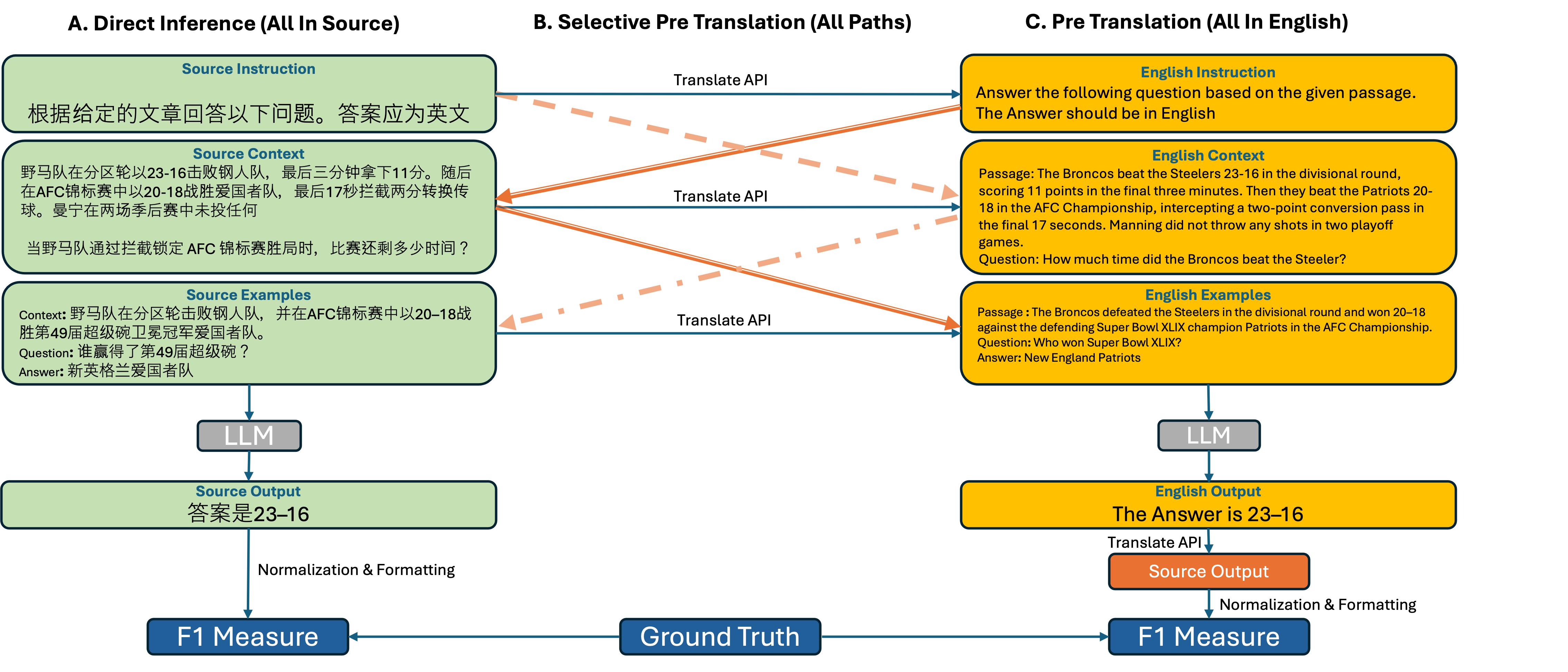}
    }
   \caption{Prompting Strategies: \emph{Direct Inference}, \emph{Selective Pre-Translation}, and \emph{Pre-Translation}}

    % Bar colors correspond to the log-normalized number of tokens in GPT-3 pre-training data.
    
    \label{fig:schema}
\end{figure*}

%Pre-Translatio, add another level between the english output to the translated output. 
%F1 measure one and both income into the same f1. 
%Remove the down arrows. 
% Add arrow for the green components. 

\section{Introduction}

Large language models (LLMs) demonstrate impressive 
%multilingual
capabilities across various natural language processing tasks, including machine translation \citep{kocmi2023findings}, natural language understanding \citep{saba2024llms} 
 %\reut{what is knowledge utilization? I am not familiar with this task}
 and complex reasoning tasks \citep{huang2022towards}. These exceptional capabilities of LLMs stem, to a large extent, from the vast amounts of data they were trained on \citep{kaplan2020scaling}. Current LLMs are primarily trained on English data but also include data from other languages, i.e., GPT-3 was trained on 119 languages, but only 7\% of the tokens are from non-English languages.\footnote{\url{https://github.com/openai/gpt3/blob/master/dataset_statistics}} With over 7,000 languages spoken worldwide \citep{anderson2010many}, the increasing pace of globalization has amplified the need for LLMs that  understand and respond in diverse languages.%%%\reut{this is a weak start, we do not only need to "prompt the LLM" in a languages, we use the LLM, which means the LLM has to have capabilities in order to repond. Try to sharpen this} \reut{in general - this is an important paragraph but it misses the mark. needs further work}

One common strategy to respond to a task presented in a language different than English
%\reut{a strategy for what?} 
is \emph{pre-translation}, which involves translating the complete prompt into English before querying the model \citep{ahuja2023mega, shi2022language}, allowing to leverage the robust capabilities acquired in  English  across different languages. %%\reut{it is not a quite multilingual task -- it is a task presented in a different language. Multilingual tasks involve several languages}
%At the same time,  previous research has shown that LLMs  as the GPT family can perform a wide range of tasks in different languages, and outperform the same task when using {\em direct inference}, i.e.,  prompts written in the source language. %\reut {monolingual in what terms? english or the source language? - sentence unclear} \citep{bareiss2024english, intrator2024breaking, chowdhery2023palm, qin2023chatgpt, ahuja2023mega}. 
%However,
At the same time, this approach introduces complexities and risks of information loss \citep{nicholas2023lost}. 
Also, it is unclear whether this approach is uniformly effective across languages and tasks, especially tasks  requiring region-specific or culturally-apropriate knowledge.
 
In contrast to pre-translation, recent studies  show that {\em direct inference}, i.e., prompting the model directly in the  (non-English) {\em source} language spoken by the user,
%%\reut{what is direct inference - you have not yet defined it with respect to which language}
outperforms pre-translation for tasks like QA \citep{intrator2024breaking}. 
However, it is unclear whether this approach is optimal, considering that the model was trained on limited data in the source language. It is also unclear how much information is shared across languages during pre-training. Be that as it may, as we show in Sec.~\ref{chap_4_compare_methods}, direct inference results still remain suboptimal.
%advantages of model output in English for cultural knowledge and generative tasks. \reut{whats the shortcoming? why not to use only that?}

% while considering more nuanced post-translation approach during inference. 

% While some works translate the prompt as a single unit \citep{bareiss2024english, intrator2024breaking}, 
In view of these shortcomings,
%\reut{you only discuss shortconings of pre-trans} 
various studies propose to use \emph{selective pre-translation}, a more nuanced method compared to the de-facto standard \emph{pre-translation} approach, which calls for translating only specific parts of the prompt \citep{ahuja2023mega, kim2023boosting, kim2024translating}. For example, \citet{liu2024translation} show that translating only the context to English outperforms direct-inference in summarization and NLI. \citet{ahuja2023mega} translated few-shot examples to English while keeping the context in the source language.
%%\reut{whats left in the source?}
\citet{kim2023boosting} used the selective approach when prompting different cross-lingual compositions of in-context examples. However, the selective approach lacks a systematic evaluation of more complex setups, e.g., instruction in English and output in the source language. Consequently, the efficacy of  \emph{selective pre-translation} and the optimal prompt configurations for various multilingual settings and tasks remain unclear. To fill this gap, in this paper we set out to examine the impact of selective pre-translation, a commonly used method, across diverse tasks, in order to devise effective prompting strategies for multilingual LLMs.

% Additionally, to our knowledge, only a handful of studies \citep{jain2019entity} effectively use translation for tasks that have no straightforward alignment between labels in translated text such as NER.\reut{why is there no alignment for NER? sounds odd} Motivated by this research gap, in our study, we examine the impact of selective pre-translation on diverse tasks with the aim of devising effective prompting strategies for multilingual LLMs.

% Specifically, in this paper we propose to formalize a configuration for a prompt ---  four functional parts: instruction, context, examples (zero/few-shot), and output --- either of which could be {\em selectively} pre-translated or not (as also stated by \citet{winata2021language, ahuja2023mega}). %We created a framework for exhaustively analyzing all

Concretely, we define a formal {\em configuration} for a prompt --- consisting of four functional parts: instruction, context, examples, and output --- either of which could be {\em selectively} pre-translated or not (see also \citet{winata2021language, ahuja2023mega}). We exhaustively assess
%%\reut{creating the framework is less important, its an  implementation details} settings of cross-lingual prompt translation
all configurations of cross-lingual prompt translation into English from different source languages.
Figure~\ref{fig:schema} presents an overview  of our approach, demonstrating the various \emph{selective pre-translation} strategies  compared against \emph{pre-translation} and \emph{direct inference} in the source language.

Through a comprehensive evaluation involving 35 languages, four tasks, six dataset collections, and three models, our results  demonstrate that \emph{selective pre-translation} consistently outperforms both \emph{pre-translation} and \emph{direct inference} in the source language, establishing the efficacy of \emph{selective pre-translation} strategies (Section~\ref{chap_4}). 
Additionally, we analyze the considerations in determining which component to translate, and illustrate the optimal strategies across tasks and languages with varying resource levels. 
Moreover, we examine how factors such as language similarity to English, training size, and language script
%%\reut{wdym by "family of data"?}
affect  task performance, and 
  show the effectiveness of selective pre-translation method in mitigating various translation issues, by choosing which prompt components to translate (Section~\ref{chap_5}).

% Additionally, we analyze how various factors, including the type of task, size and family of pre-trained data, language similarity to English, impact the performance of our proposed {\em selective} approach. Furthermore, based on these factors, we provide guidelines for implementing optimal pre-translation strategies. Finally, we perform an additional experiment, analyzing the impact of translation on \emph{pre-translation}. 

More specifically, our findings   demonstrate that in extractive tasks such as QA or NER, where the output overlaps with the provided context and no generation is needed, the model is either agnostic to the context language in the case of high-resource languages or prefers  context in the source language in the case of low-resource languages. Surprisingly, we have discovered that low-resource languages yield better results even when the model's output is required in English, e.g., in NER (Section~\ref{chap_4}). %%\reut{again - why no alignment?} 
 Moreover,  we show that translation quality significantly affects model performance and that the \emph{selective pre-translation} approach essentially mitigates the negative effect of suboptimal translations, which are in turn specifically problematic in lower-resourced languages (Section~\ref{chap_5}).
 %\reut{please verify that this last sentence is correct, and if not remove it}

%By systematically analyzing the impact of selectively translating each component, we can provide a more nuanced understanding of cross-lingual pre-translation, which generalizes 
All in all, our extensive  and  systematic evaluation of pre-translation strategies facilitates generalization across a broader range of languages and tasks, beyond  the specific ones herein, towards more robust LLM-use in multilingual settings.

\section{The Proposal: Formalizing Prompts  {\em Selective Pre-Translation}  Strategies}

Current practices of {\em prompting}  generative LLMs, such as the GPT models family %-3.5-turbo
\citep{ouyang2022training} and Gemini \citep{team2023gemini}, %%\reut{you need to qualify which LLMs you mean -- bc definitely not all LLMs perform instruction following eg BERT ROBERTA and T5 -- so which models are your target?}
uncover two remarkable capabilities in  performing language processing tasks: (i) \textit{chain of thought} \citep{wei2022chain}, where LLMs solve complex tasks through a series of intermediate reasoning steps, and (ii) \textit{in-context learning} \citep{brown2020language}, allowing the model to adapt to new tasks based on limited examples, without weights updates. These capabilities are built on top of the notion of a {\em prompt}, which serves as a prefix for the LLM's response. 
These capabilities are powered by  the complex  nuanced structure of nowadays prompts, consisting of 
%To harness these capabilities, our approach formalizes the prompt into
four components: {\em instruction, context, examples}, and {\em output}.

% %We define four components that constitute the prompt. 
% We refine \citet{ahuja2023mega}'s definition of prompt components to include these four components. 

%Specifically, we define them as 
Let us first define these four components, as
follows. The \textbf{\em Instruction} (\(I\)) provides a natural language guidance to the model, explaining the task to be performed. The \textbf{\em Context} (\(X\)) represents the task data that the model operates on in performing the task. \textbf{\em Examples} (\(E\)) are optional illustrations of context:output pairs, that can be used for in-context learning.
Overall, we define \(\langle I  \, X\,\rangle\) as a zero-shot prompt and a few-shot   prompt  as  \(\langle I  \, X  \, E\,\rangle\).
 The prompt is processed by a model \(M\) to yield an \textbf{\em Output} (\(O\)), where the instruction can include a request for the model to generate the output in a specific language
or format.

%\reut{this is NOT a good sign - it denotes intersection. why not using \_ or simply $$ as connectors? You can also use the $\hat{}$ sign to concatenate} 

%\reut{Actually, this formalization is somewhat lacking because E is essentially a sequence of X:O correct pairs. could be better to define zero shot with IXO, first and then define E and the add fewshot IXEO}
Each component, i.e., the instruction, the context,  the examples, or the output, may be pre-translated or not. We denote a pre-translation decision  \( l \in \{ e, s \} \), where  e stands for  English pre-translation and s for the source language. Standardly the prompt is  composed as  \(\langle I  \, X \, E\,\rangle\) and is delivered to the model M, which in turn emits an output \(O\). We define a specific {\em  pre-translation configuration}\footnote{See Appendix \ref{appendix:configuration_format} for specific configuration examples.}   as 
\( c = \langle \text{I}^{l_i}, \text{X}^{l_x}, E^{l_e}, O^{l_o} \rangle \)
% \( c = \langle {l_I}, {l_X}, {l_E}, {l_O} \rangle \)
where the subscript \( l \in \{ e, s \} \) indicates the language of the component. Having defined the pre-translation configurations, we evaluate them in different settings.

%%\reut{I think that there was no point where you specifically said you are overall interested in a task that is completely completed in a language different than English, and English is just a helper, Bc some ppl ask in English about multiling stuff and still want the result in English. Thats quite confusing to the average reader.}

% Each component, including the instruction, context, and examples (I, X, and E), can be pre-translated, denoted as \(Component^L\) where L can be English or the source language. The instruction can include a request for the model to generate an output in a specific language or format. Overall, each configuration can be denoted as a set of four components: \(\langle Instruction^L, Input^L, Examples^L, Output^L \rangle\), where \(L\) can be English or source language. 

\section{Selective Pre-Translation Evaluation}
\label{chap_4}

% In this section, we evaluate various selective pre-translation strategies and assess the configuration selection impact on model performance.

\subsection{Experimental Setup}

\paragraph{Goal}
We set out to compare {\em selective pre-translation} to both {\em pre-translation} and {\em direct inference}, and to assess the impact of the selected configuration on task performance across languages.

\paragraph{Prompt Configuration}
We assess \emph{selective pre-translation} in both zero-shot and few-shot settings.  In the zero-shot settings, with no examples, we considered \(2^3\) configurations. For the few-shot scenario, with four components, each is either translated to English or retains the source language, we get \(2^4\) configurations. %\reut{so how many is this? calculate}.
All in all  we experiments with \(24\) configurations per language and task.\footnote{NLI has 12 configurations, with output always in the instruction language, due to its particular, fixed, output format.}
%spelled-out index-based format.}
%\reut{why? you could translate the labels as well. need a better justification}

%\reut{for both 0 and few shot? show the calc} 

\paragraph{Prompt Creation And Output Normalization} 
Based on the prompt configuration, we used the Google Translate API\footnote{\url{pypi.org/project/easygoogletranslate/}} to translate the components that required translation. After querying the model, we normalized and formatted its output, then translated it to match the language of the gold standard. See Appendix \ref{appendix:modular_prompting_experimental_setup} for further implementation.

% \begin{itemize}
%     \item Few shot
%     \item Zero shot
% \end{itemize}
%  We systematically alter the language (English/Source) of each component: instruction, context, examples, and output. We examine 24 configurations per task, 16 ((English/Source)$^4$) for few-shot and 8 ((English/Source)$^3$) for zero-shot, except for NLI where the output is in English.

\paragraph{Models}

We conducted experiments on several LLMs: (1) Standard generative models—GPT-3.5-turbo \citep{ouyang2022training}, Mistral-8x7B \citep{jiang2023mistral}, and Gemini-1.0-pro \citep{team2023gemini} — with context sizes of 16k, 32k, and 8k, respectively; (2) multilingual - bloomz-7b1-mt \citep{muennighoff2022crosslingual}, with a 2k context.\footnote{ See Appendix \ref{appendix:models_platforms} for details on the models we used.}

\begin{table}[tb]
\resizebox{\columnwidth}{!}{%
\begin{tabular}{lllll}
\multicolumn{1}{c}{\textit{\textbf{Affinity}}}       & \textbf{Class}                  & \multicolumn{1}{c}{\textit{\textbf{Range (\% of tokens)}}} & \textit{\textbf{Avg. \#tokens (M)}} & \multicolumn{1}{c}{\textit{\textbf{STD}}} \\ \hline
\multicolumn{1}{l|}{High Resource}           & \multicolumn{1}{l|}{A} & \multicolumn{1}{l|}{\textit{p} $\geq$ 0.1\%}              & 1,240                                & 1,156                                   \\ \hline
\multicolumn{1}{l|}{Medium Resource}         & \multicolumn{1}{l|}{B} & \multicolumn{1}{l|}{0.01\% $<$ \textit{p} $<$ 0.1\%}      & \pz\pz72                             & \pz\pz49                                \\ \hline
\multicolumn{1}{l|}{Low Resource}            & \multicolumn{1}{l|}{C} & \multicolumn{1}{l|}{0\% $<$ \textit{p} $\leq$ 0.01\%}     & \pz\pz\pz5.07                        & \pz\pz\pz5.41                            \\ \hline
\multicolumn{1}{l|}{Unrepresented}           & \multicolumn{1}{l|}{D} & \multicolumn{1}{l|}{\textit{p} = 0\%}                     & \pz\pz\pz0                           & \pz\pz\pz0                               \\ \hline
\end{tabular}%
}
\caption{Language categorization  based on the percentage (\(p\)) of tokens per language in GPT-3's training data. Avg. token count (millions), \(\text{STD}\): standard deviation.}
\label{tab:classes_summary}
\end{table}

\paragraph{Language Selection and Categorization}
We selected \textasciitilde 11 languages per task, ensuring a balanced representation across resource levels (high, medium, low). Due to the lack of precise pre-training distribution for the LLMs we use, we employed the GPT-3 distribution as a proxy, as it is the only distribution publicly shared, to our knowledge.\footnote{\url{https://github.com/openai/gpt-3/blob/master/dataset_statistics}}
%\reut{put link / ref where the rev can find this distribution}
The  GPT-3's  multilingual coverage
%\reut{didnt we just say before that gpt models are english centric?}
enables us to categorize languages into classes based on their data ratios. Following \citet{lai2023chatgpt}, we categorized the tested languages into four classes based on data ratio:
%\reut{could the categorize vary based on the model? say, chinese, can be high in GPT and low in Mixtrall for instance? where is the line? how do you justify it? consider adding an explanation of a footnote} 
High-Resource (A), Medium-Resource (B), Low-Resource (C), and Unrepresented (D).\footnote{See Table \ref{tab:languages_classes} for list of languages, codes and data ratios.} Class D, which includes languages unseen during training. Table \ref{tab:classes_summary} summarizes this classification with basic properties.\footnote{Alternative criteria such as speakers ratio, as proposed by \citet{joshi2020state}, do not  reflect language diversity in LLMs, which is affected by  availability of data rather than speakers.}

\begin{table}[tb]
\centering
\resizebox{\columnwidth}{!}{%
\begin{tabular}{lll}
\multicolumn{1}{c}{\textit{\textbf{Task}}} & \multicolumn{1}{c}{\textit{\textbf{Dataset}}} & \multicolumn{1}{c}{\textit{\textbf{Languages}}}                                                                                                                               \\ \hline
\multicolumn{1}{l|}{NLI}                   & \multicolumn{1}{l|}{XNLI}                     & \begin{tabular}[c]{@{}l@{}}\textit{\textbf{High}}: Spanish, German, Chinese \\
\textit{\textbf{Medium}}: Greek, Turkuish, Arabic \\ \textit{\textbf{Low}}: Bulgarian,  Hindi,  Thai, Swahili, Urdu\\ \end{tabular} \\ \hline
\multicolumn{1}{l|}{\multirow{2}{*}{QA}}   & \multicolumn{1}{l|}{XQuAD}                    & \begin{tabular}[c]{@{}l@{}}\textbf{\textit{High}}: German, Russian, Romanian \\ \textbf{\textit{Medium}}: Arabic,  Greek,  Vietnamese\end{tabular}                                                                 \\ \cline{2-3} 
\multicolumn{1}{l|}{}                      & \multicolumn{1}{l|}{IndicQA}                  & \begin{tabular}[c]{@{}l@{}}\textbf{\textit{Low}}:  Hindi,  Malayalam, Bengali,  Telugu \\ \textbf{\textit{Unrepresented}}:  Assamese \end{tabular}                                                          \\ \hline
\multicolumn{1}{l|}{\multirow{2}{*}{NER}}  & \multicolumn{1}{l|}{MasakhaNER}               & \textbf{\textit{Unrepresented}}: Bambara, Ese, Hausa, Yoruba                                                                                                                                    \\ \cline{2-3} 
\multicolumn{1}{l|}{}                      & \multicolumn{1}{l|}{WikiANN}                  & \begin{tabular}[c]{@{}l@{}}\textbf{\textit{High}}: French, Chinese, Italian, Portuguese,  Swedish\\ \textbf{\textit{Medium}}: Serbian, Slovak\end{tabular}                                                        \\ \hline
\multicolumn{1}{l|}{Summarization}         & \multicolumn{1}{l|}{XL-Sum}                   & \begin{tabular}[c]{@{}l@{}}\textbf{\textit{High}}:  French, Japanese,  Spanish,  Portuguese\\ \textbf{\textit{Medium}}:  Korean, Turkish, \\ \textbf{\textit{Low}}:  Azerbaijani, Nepali,  Persian, Uzbek\end{tabular}              \\ \hline
\end{tabular}%
}
\caption{Experiment Setup: Tasks, datasets, languages. Languages are separated by their resource-type affinity.}
\label{tab:tasks_datasets}
\end{table}

\begin{table*}[tb]
\centering
\resizebox{\textwidth}{!}{
\begin{tabular}{cccccccccccccccccccc}
\hline
\multicolumn{5}{c}{\textbf{Question Answering (QA, F1)}}                                                                                                                               & \multicolumn{5}{c}{\textbf{Summarization (ROUGE)}}                                                                                                                                                                                                                     & \multicolumn{5}{c}{\textbf{Named Entity Recognition (NER, F1)}}                                                                                                                                                                                                        & \multicolumn{5}{c}{\textbf{Natural Language Inference (NLI, Acc.)}}                                                                                                                                                                                                    \\
\hline
\textit{Lng} & \textit{Cls.} & \(\uparrow\)\textit{Top} & \textit{Src.} (\%) & \textit{Eng.} (\%) & 
\textit{Lng} & \textit{Cls.} & \(\uparrow\)\textit{Top} & \textit{Src.} (\%) & \textit{Eng.} (\%) &
\textit{Lng} & \textit{Cls.} & \(\uparrow\)\textit{Top} & \textit{Src.} (\%) & \textit{Eng.} (\%) &
\textit{Lng} & \textit{Cls.} & \(\uparrow\)\textit{Top} & \textit{Src.} (\%) & \textit{Eng.} (\%) \\ 
\hline
en           & A            & 0.77                            & N.A                                             & \multicolumn{1}{c|}{N.A}                                            & en                               & A                                & 30.23                                               & N.A                                                                 & \multicolumn{1}{c|}{N.A}                                            & en                               & A                                & 0.65                                                & N.A                                                                 & \multicolumn{1}{c|}{N.A}                                            & en                               & A                                & 0.69                                                & N.A                                                                 & N.A                                                                 \\ \hline
de           & A             & 0.85                            & \pz18\%                         & \multicolumn{1}{c|}{\pz\pz9\%}       & fr                               & A                                 & 35.12                                               & 16\%                                                               & \multicolumn{1}{c|}{10\%}                                          & sr                               & B                                 & 0.77                                                & 52\%                                                               & \multicolumn{1}{c|}{265\%}                                         & sw                               & C                                 & 0.73                                                & 58\%                                                               & 28\%                                                               \\
hi           & C             & 0.82                            & \pz32\%                         & \multicolumn{1}{c|}{182\%}                                         & ja                               & A                                 & 32.47                                               & 17\%                                                               & \multicolumn{1}{c|}{14\%}                                          & it                               & A                                 & 0.75                                                & \pz9\%                                              & \multicolumn{1}{c|}{\pz41\%}                        & bg                               & C                                 & 0.72                                                & 57\%                                                               & \pz8\%                                              \\
ar           & B             & 0.74                            & \pz84\%                         & \multicolumn{1}{c|}{138\%}                                         & fa                               & C                                 & 29.34                                               & 21\%                                                               & \multicolumn{1}{c|}{\pz0\%}                         & sk                               & B                                 & 0.72                                                & 15\%                                                               & \multicolumn{1}{c|}{\pz36\%}                        & el                               & B                                 & 0.71                                                & 24\%                                                               & 30\%                                                               \\
vi           & B             & 0.73                            & \pz\pz0\%        & \multicolumn{1}{c|}{\pz58\%}                        & es                               & A                                 & 28.28                                               & 10\%                                                               & \multicolumn{1}{c|}{\pz3\%}                         & po                               & A                                 & 0.72                                                & 18\%                                                               & \multicolumn{1}{c|}{\pz20\%}                        & es                               & A                                 & 0.69                                                & 20\%                                                               & 18\%                                                               \\
ro           & A             & 0.69                            & \pz\pz0\%        & \multicolumn{1}{c|}{\pz\pz9\%}       & po                               & A                                 & 27.40                                               & \pz8\%                                              & \multicolumn{1}{c|}{\pz0\%}                         & fr                               & A                                 & 0.72                                                & 23\%                                                               & \multicolumn{1}{c|}{\pz24\%}                        & ar                               & B                                 & 0.67                                                & 28\%                                                               & 23\%                                                               \\
ru           & A             & 0.69                            & \pz\pz6\%        & \multicolumn{1}{c|}{305\%}                                         & tr                               & B                                 & 20.87                                               & 18\%                                                               & \multicolumn{1}{c|}{\pz0\%}                         & hau                              & C                                 & 0.70                                                & 62\%                                                               & \multicolumn{1}{c|}{\pz51\%}                        & hi                               & C                                 & 0.64                                                & 59\%                                                               & \pz8\%                                              \\
el           & B             & 0.69                            & \pz\pz0\%        & \multicolumn{1}{c|}{\pz\pz2\%}       & ne                               & C                                 & 19.58                                               & 31\%                                                               & \multicolumn{1}{c|}{28\%}                                          & ee                               & D                                 & 0.68                                                & 46\%                                                               & \multicolumn{1}{c|}{\pz81\%}                        & de                               & A                                 & 0.64                                                & 19\%                                                               & \pz8\%                                              \\
bn           & D             & 0.68                            & \pz44\%                         & \multicolumn{1}{c|}{423\%}                                         & as                               & D                                 & 15.79                                               & 17\%                                                               & \multicolumn{1}{c|}{\pz7\%}                         & sv                               & A                                 & 0.68                                                & 12\%                                                               & \multicolumn{1}{c|}{\pz\pz9\%}       & zh                               & B                                 & 0.63                                                & 16\%                                                               & \pz4\%                                              \\
as           & D             & 0.56                            & 138\%                                          & \multicolumn{1}{c|}{450\%}                                         & uz                               & C                                 & 15.72                                               & 58\%                                                               & \multicolumn{1}{c|}{24\%}                                          & zh                               & B                                 & 0.63                                                & 90\%                                                               & \multicolumn{1}{c|}{121\%}                                         & th                               & B                                 & 0.57                                                & 49\%                                                               & 10\%                                                               \\
te           & C             & 0.53                            & 210.10\%                                       & \multicolumn{1}{c|}{253.30\%}                                      & ko                               & B                                 & 11.84                                               & 36.99\%                                                            & \multicolumn{1}{c|}{11.78\%}                                       & bam                              & D                                 & 0.33                                                & 33.25\%                                                            & \multicolumn{1}{c|}{\pz80.24\%}                     & ur                               & C                                 & 0.57                                                & 29.96\%                                                            & \pz9.08\%                                           \\
ml           & C             & 0.49                            & 104.30\%                                       & \multicolumn{1}{c|}{600.00\%}                                      &                                  &                                   &                                                     &                                                                     & \multicolumn{1}{c|}{}                                               & yor                              & D                                 & 0.32                                                & 66.02\%                                                            & \multicolumn{1}{c|}{\pz49.19\%}                     & \multicolumn{1}{l}{tr}           & \multicolumn{1}{l}{B}             & \multicolumn{1}{l}{0.57}                            & \pz0.00\%                                           & \pz8.14\%                                           \\ \hline
\end{tabular}%
}
\caption{For each Language, we present the top-performing selective configuration score over all other configurations ({\it Top}) along with its relative improvement (\%) over \emph{direct inference} (\textit{Src}) and {\em  pre-translation} (Eng).} 
%, percentage improvement over \emph{few-shot pre-translation} (\textit{Eng.}), and the language class (as detailed in Table \ref{tab:languages_classes}).}}

\label{tab:summary_results}
\end{table*}

\paragraph{Tasks and Datasets}

We assess model performance on 4 tasks, NLI, QA, NER, and Summarization, which we detail in turn. (i)
%%\reut{This is unclear - you experiment with all languages in each of these dataset? bc the table above shows a 10 for each. if you selected then how did you sample the languages? did you care for the overlap (that all languages will have all tasks?)}
\textbf{\emph{Natural Language Inference (NLI)}} determines if a hypothesis entails, contradicts, or is neutral to a premise. We use the XNLI dataset \citep{conneau-etal-2018-xnli}, with sentence pairs in 11 languages, and measured  prediction accuracy.
% by comparing model predictions to ground truth labels.
% \paragraph{Natural Language Inference (NLI)} NLI involves determining if a hypothesis is entailed by, contradicts, or is neutral to a premise. We evaluated this using the XNLI dataset \citep{conneau-etal-2018-xnli}, which contains premise-hypothesis pairs in 15 languages. The input is a pair of sentences, and the output is a classification label: entailment, contradiction, or neutral. We measure performance using accuracy, comparing the model's predictions to the dataset's ground truth labels.
% \paragraph{Question Answering (QA)} We focus on extractive question answering, where the model must identify the relevant answer span within a given context.  We evaluate the models on two datasets: XQuAD \citep{Artetxe:etal:2019} and IndicQA for Indic languages \citep{doddapaneni2022indicxtreme}. Model performance was assessed based on the exact match between predicted and ground truth answer spans, using the F1 metric.
(ii) \emph{\textbf{Question Answering (QA):}} We focus on extractive QA, where the model identifies the answer span in a given context. We evaluated performance on XQuAD \citep{Artetxe:etal:2019} and IndicQA \citep{doddapaneni2022indicxtreme} for Indic languages, using the F1 score to assess performance.
(iii) \emph{\textbf{Named Entity Recognition (NER):}}
% In this task, the model is instructed to identify and classify named entities within a given sentence. We employed two datasets for evaluation: WikiANN \citep{pan2017cross} and MasakhaNER \citep{adelani2021masakhaner}. WikiANN is a widely used NER dataset containing Wikipedia sentences annotated with LOC (Location), PER (Person), and ORG (organization) tags which supports 176 languages. MasakhaNER is specifically designed for African languages. Both datasets utilize the BIOSE scheme (B=Begin, I=Inside, O=Outside, S=Singleton, E=End) to mark entity boundaries. However, for our experiments, we rephrased the task as a generative task,  instructing the model to directly output entity spans without requiring strict adherence to the BIOSE tags. We evaluated the model using the F1 metric.
% The task of identifying named entities in sentences.
We sampled languages from %\reut{sampled language from? you didnt evaluate on ALL} 
two datasets: WikiANN \citep{pan2017cross}, which includes Wikipedia sentences annotated with Location, Person, and Organization tags in 176 languages; and MasakhaNER \citep{adelani2021masakhaner}, for African languages. While both datasets use the BIOSE scheme to delineate entity boundaries, we recast the task as generative, prompting the model to generate the list labeled named entities for a given input context. Model performance has been evaluated using  F1  scores.
% \paragraph{Abstractive Text Summarization} Abstractive summarization is the task where the model generates concise and informative summaries from longer texts by generating new text, unlike extractive summarization, which selects existing sentences. We evaluated our model using the XL-Sum dataset \citep{hasan2021xl}, which provides summaries of news articles in 44 languages, making it ideal for evaluating multilingual summarization models. We used the ROUGE metric for evaluation, which measures the overlap between generated summaries and reference summaries.
(iv)~\emph{\textbf{Abstractive Text Summarization}} involves generating  short summaries of long contexts, rather than extracting existing sentences.
% from longer texts, as opposed to extractive summarization, which selects existing sentences.
We used the XL-Sum dataset \citep{hasan2021xl}, which offers news article summaries in diverse languages. We sampled 10 languages from the dataset and
% , suitable for multilingual evaluation. 
evaluated with ROUGE.
We conducted experiments on a sample of 250 examples\footnote{We selected 250 that followed the can fit into the context of the model, i.e., \(<16K\).} 
%It was not chosen randomly because the large-context articles in XL-Sum are computationally expensive.
 from the test sets for each language.\footnote{For tasks without public test sets (XQUAD, IndicQA), we used the validation data.}%\reut{again it sounds like we execute experiments on 44 langs. To me it seems that this is not the case - quite misleading, be precise}
In total, the datasets we use encompass 35  languages across 4 tasks.
Table \ref{tab:tasks_datasets} lists the datasets used, covering \textasciitilde 11 languages per task, ensuring a balanced representation of \{Low,Mid,High\} resource categories for each task. 

\paragraph{Analysis Methods}

To analyze the empirical results and detect the most influential components,
we use three methods: (i)~\textit{Correlation analysis} -- Assessing the relationship between the model's prediction scores and the language selection per component.
(ii)~\textit{Association Rule Learning (ARL)} and {\em the Apriori algorithm} --  While correlation analysis provides a preliminary understanding of the relationship between individual components and model performance, it does not capture non-linear relationships, i.e., the combined effect of multiple translation decisions on performance.
% or reveal unexpected configurations that might be crucial for model performance. 
To address this limitation, we utilize ARL with the Apriori algorithm \citep{piatetsky1991discovery}.\footnote{Appendix  
\ref{appendix:rule_association_recap_implementations} details the algorithm and implementation.} 
(iii) ~\textit{Performance Gap} --  We computed the average difference of k configuration pairs \( c_{i}\) and \( c_{j}\) such that they differ only in the language of one component, e.g., 
\(\langle I^{\textbf{e}}, X^e, E^{e}, O^e \rangle \)
%\(\langle e^{\textbf{I}}, e^X, e^{E}, e^O \rangle \) and \(\langle s^{\textbf{I}}, e^X, e^{E}, e^O \rangle \).
%
and \(\langle I^{\textbf{s}}, X^e, E^{e}, O^e \rangle \).
We then calculated this average to determine the  performance gap for each specific task: \(\frac{1}{k} \sum_{\langle i,j\rangle =1}^{k} \left( \text{Eval}(c_i) - \text{Eval}(c_j) \right)\), where \(\text{Eval}()\) denotes the task evaluation  score, and \(k\) is the number of distinct pairs.
%\reut{for a specific task? for all tasks? unclear what is the task status in this calculation }

\begin{table*}[]
\centering
\scalebox{0.995}{
\resizebox{\textwidth}{!}{
\begin{tabular}{lllllllllllllllllllllll}
\multicolumn{5}{c}{\underline{QA}}                                                                                          &                                              & \multicolumn{5}{c}{\underline{Summarization}}                                                                               &                                              & \multicolumn{5}{c}{\underline{NER}}                                                                                         &                                              & \multicolumn{4}{c}{\underline{NLI}}                                                              &                         \\
\textit{\textbf{lng}} & \textit{\textbf{cls.}} & \textit{\textbf{instruction}} & \textit{\textbf{context}} & \textit{\textbf{examples}} & \multicolumn{1}{l|}{\textit{\textbf{output}}} & \textit{\textbf{lng}} & \textit{\textbf{cls.}} & \textit{\textbf{instruction}} & \textit{\textbf{context}} & \textit{\textbf{examples}} & \multicolumn{1}{l|}{\textit{\textbf{output}}} & \textit{\textbf{lng}} & \textit{\textbf{cls.}} & \textit{\textbf{instruction}} & \textit{\textbf{context}} & \textit{\textbf{examples}} & \multicolumn{1}{l|}{\textit{\textbf{output}}} & \textit{\textbf{lng}} & \textit{\textbf{cls.}} & \textit{\textbf{instruction}} & \textit{\textbf{context}} & \textit{\textbf{examples}} \\
ru                    & A                      & -0.08**                       & \textbf{0.35**}           & 0.12**                     & \multicolumn{1}{l|}{0.09**}                   & ja                    & A                      & \textbf{-0.33**}              & -0.08                     & -0.02*                     & \multicolumn{1}{l|}{0.00}                     & fr                    & A                      & -0.11*                        & 0.10*                     & -0.01                      & \multicolumn{1}{l|}{0.01}                     & de                    & A                      & -0.03                         & -0.02                     & -0.01                      \\
de                    & A                      & -0.03**                       & \textbf{0.30**}           & 0.08                       & \multicolumn{1}{l|}{0.03*}                    & fr                    & A                      & \pz0.01                       & 0.020                     & -0.04                      & \multicolumn{1}{l|}{0.06}                     & it                    & A                      & \pz0.02                       & 0.04                      & -0.04                      & \multicolumn{1}{l|}{0.01}                     & es                    & A                      & -0.03                         & 0.02                      & -0.03*                     \\
ro                    & A                      & -0.03                         & 0.12**                    & 0.04                       & \multicolumn{1}{l|}{0.02}                     & po                    & A                      & -0.08*                        & 0.05*                     & -0.03*                     & \multicolumn{1}{l|}{0.10*}                    & po                    & A                      & -0.15                         & 0.09*                     & \pz0.1                     & \multicolumn{1}{l|}{0.01}                     & el                    & B                      & -0.04                         & 0.01                      & 0.07                       \\
vi                    & B                      & \pz0.04                       & \textbf{0.40**}           & 0.10**                     & \multicolumn{1}{l|}{0.10}                     & es                    & A                      & -0.09*                        & 0.03*                     & -0.03                      & \multicolumn{1}{l|}{0.05}                     & sv                    & A                      & -0.11*                        & 0.06*                     & -0.03**                    & \multicolumn{1}{l|}{0.01}                     & zh                    & B                      & \pz0.01                       & -0.06                     & -0.06                      \\
ar                    & B                      & -0.07**                       & \textbf{0.20**}           & 0.13**                     & \multicolumn{1}{l|}{0.04*}                    & tr                    & B                      & -0.14**                       & 0.10                      & -0.1                       & \multicolumn{1}{l|}{-0.03*}                   & zh                    & B                      & -0.26**                       & \textbf{0.44**}           & \pz0.00                    & \multicolumn{1}{l|}{0.07}                     & ar                    & B                      & \pz0.00                       & -0.02                     & -0.04                      \\
el                    & B                      & -0.06                         & \textbf{0.48**}           & 0.03                       & \multicolumn{1}{l|}{0.07*}                    & ko                    & B                      & -0.10**                       & 0.13                      & 0.01                       & \multicolumn{1}{l|}{0.05}                     & sr                    & B                      & -0.26**                       & \textbf{0.44**}           & \pz0.09**                  & \multicolumn{1}{l|}{0.05}                     & th                    & B                      & -0.03                         & 0.03                      & -0.14*                     \\
bn                    & C                      & -0.10**                       & \textbf{0.38**}           & 0.03                       & \multicolumn{1}{l|}{0.03}                     & uz                    & C                      & \textbf{-0.42**}              & 0.14                      & 0.03                       & \multicolumn{1}{l|}{-0.12*}                   & sk                    & B                      & -0.11**                       & \textbf{0.30**}           & -0.1*                      & \multicolumn{1}{l|}{0.01}                     & tr                    & B                      & -0.02                         & 0.00                      & 0.02                       \\
ma                    & C                      & -0.14**                       & \textbf{0.30**}           & 0.01                       & \multicolumn{1}{l|}{0.03}                     & fa                    & C                      & \textbf{-0.37**}              & 0.05                      & -0.07**                    & \multicolumn{1}{l|}{-0.04}                    & bam                   & D                      & \pz0.03                       & \textbf{0.44**}           & -0.11                      & \multicolumn{1}{l|}{0.02}                     & ur                    & C                      & \pz0.01                       & 0.01                      & -0.08*                     \\
te                    & C                      & -0.10**                       & \textbf{0.38**}           & 0.03                       & \multicolumn{1}{l|}{0.03}                     & ne                    & C                      & \textbf{-0.35**}              & -0.09                     & 0.07**                     & \multicolumn{1}{l|}{-0.14}                    & ewe                   & D                      & -0.01                         & \textbf{0.38**}           & -0.12**                    & \multicolumn{1}{l|}{0.01}                     & bg                    & C                      & \pz0.01                       & 0.05                      & -0.13*                     \\
hi                    & C                      & -0.07**                       & \textbf{0.30**}           & 0.05                       & \multicolumn{1}{l|}{0.01}                     & az                    & C                      & \textbf{-0.30**}              & 0.04                      & -0.00                      & \multicolumn{1}{l|}{-0.05}                    & yo                    & D                      & -0.01                         & \textbf{0.36**}           & \pz0.01*                   & \multicolumn{1}{l|}{0.03}                     & sw                    & C                      & \pz0.12                       & -0.06                     & -0.09                      \\
as                    & D                      & -0.04**                       & \textbf{0.30**}           & 0.06                       & \multicolumn{1}{l|}{0.06}                     &                       &                        &                               &                           &                            & \multicolumn{1}{l|}{}                         & hau                   & D                      & -0.04                         & \textbf{0.30**}           & \pz0.08*                   & \multicolumn{1}{l|}{0.02}                     & hi                    & C                      & -0.03                         & -0.09                     & -0.09**                    \\ \hline
\end{tabular}%
}
}

\caption{For each language, we present the Point-biserial correlation ($\tau$) between the individual component's language selection (English/Source), and the model performance score across all the configurations samples that use it. 
% Pearson correlation ($\tau$) between GPT-3.5 Turbo performance and language selection for each component.
% by task/language (lng.): \textit{instruction} \textit{context}, \textit{examples}, and \textit{output} against performance metrics. 
Positive $|\tau|$ values correlate with the source language, and negative $|\tau|$ values correlate with English. Significant correlations are indicated by *p < 0.05 and **p < 0.01. Bold values denote correlations ($|\tau| > 0.3, p < 0.01$).}
\label{tab:correlation}
\end{table*}

\begin{table}[]
\scalebox{1.0}{
\resizebox{\columnwidth}{!}{%
\begin{tabular}{clllllllclllllclllllclll}

\multicolumn{1}{l}{\textbf{Resource}}       & \textbf{Model} &                    & \multicolumn{6}{c}{\textbf{QA}}                                                                                                & \multicolumn{6}{c}{\textbf{NER}}                                                                                                                   & \multicolumn{7}{c}{\textbf{Summarization}}                                                                                                         & \multicolumn{2}{c}{\textbf{NLI}}                                                                    \\ \cline{1-2} \cline{4-24} 
\multicolumn{2}{c}{\textbf{Component}}                       &                    & I.                       & X.                       & E.                       & O.                       & \multicolumn{2}{l}{}                   & I.                       & X.                       & E.                       & O.                       & \multicolumn{2}{l}{}                   & I.                       & X.                       & E.                       & O.                       & \multicolumn{2}{l}{}                   & I.                       & X.                       & {\color[HTML]{000000} E.} \\ \cline{1-2} \cline{4-7} \cline{10-13} \cline{16-19} \cline{22-24} 
\multicolumn{1}{c|}{}                       & GPT            &                    & {\color[HTML]{3531FF} N} & {\color[HTML]{32CB00} S} & {\color[HTML]{32CB00} S} & {\color[HTML]{32CB00} S} & \multicolumn{2}{l}{}                   & {\color[HTML]{3531FF} N} & {\color[HTML]{32CB00} S} & {\color[HTML]{32CB00} S} & {\color[HTML]{32CB00} S} & \multicolumn{2}{l}{}                   & {\color[HTML]{32CB00} S} & {\color[HTML]{32CB00} S} & {\color[HTML]{3531FF} N} & {\color[HTML]{3531FF} N} & \multicolumn{2}{l}{}                   & {\color[HTML]{3531FF} N} & {\color[HTML]{32CB00} S} & {\color[HTML]{FE0000} E}  \\
\multicolumn{1}{c|}{}                       & Gemini         &                    & {\color[HTML]{009901} S} & {\color[HTML]{32CB00} S} & {\color[HTML]{32CB00} S} & {\color[HTML]{32CB00} S} & \multicolumn{2}{l}{}                   & {\color[HTML]{3531FF} N} & {\color[HTML]{32CB00} S} & {\color[HTML]{32CB00} S} & {\color[HTML]{32CB00} S} & \multicolumn{2}{l}{}                   & {\color[HTML]{FE0000} E} & {\color[HTML]{FE0000} E} & {\color[HTML]{1F1F1F} Z} & {\color[HTML]{3531FF} N} & \multicolumn{2}{l}{}                   & {\color[HTML]{3531FF} N} & {\color[HTML]{3531FF} N} & {\color[HTML]{FE0000} E}  \\
\multicolumn{1}{c|}{\multirow{-3}{*}{High}} & Mixtral        &                    & {\color[HTML]{3531FF} N} & {\color[HTML]{32CB00} S} & {\color[HTML]{32CB00} S} & {\color[HTML]{32CB00} S} & \multicolumn{2}{l}{}                   & {\color[HTML]{3531FF} N} & {\color[HTML]{32CB00} S} & {\color[HTML]{32CB00} S} & {\color[HTML]{32CB00} S} & \multicolumn{2}{l}{}                   & {\color[HTML]{32CB00} S} & {\color[HTML]{32CB00} S} & Z                        & {\color[HTML]{32CB00} S} & \multicolumn{2}{l}{}                   & {\color[HTML]{3531FF} N} & {\color[HTML]{32CB00} S} & {\color[HTML]{32CB00} S}  \\ \cline{1-2} \cline{4-7} \cline{10-13} \cline{16-19} \cline{22-24} 
\multicolumn{1}{c|}{}                       & GPT            &                    & {\color[HTML]{3531FF} N} & {\color[HTML]{32CB00} S} & {\color[HTML]{32CB00} S} & {\color[HTML]{32CB00} S} & \multicolumn{2}{l}{}                   & {\color[HTML]{3531FF} N} & {\color[HTML]{32CB00} S} & {\color[HTML]{32CB00} S} & {\color[HTML]{FE0000} E} & \multicolumn{2}{l}{}                   & {\color[HTML]{FE0000} E} & {\color[HTML]{FE0000} E} & {\color[HTML]{32CB00} S} & {\color[HTML]{FE0000} E} & \multicolumn{2}{l}{}                   & {\color[HTML]{3531FF} N} & {\color[HTML]{FE0000} E} & {\color[HTML]{32CB00} S}  \\
\multicolumn{1}{c|}{}                       & Gemini         &                    & {\color[HTML]{32CB00} S} & {\color[HTML]{32CB00} S} & {\color[HTML]{32CB00} S} & {\color[HTML]{32CB00} S} & \multicolumn{2}{l}{}                   & {\color[HTML]{FE0000} E} & {\color[HTML]{32CB00} S} & {\color[HTML]{32CB00} S} & {\color[HTML]{FE0000} E} & \multicolumn{2}{l}{}                   & {\color[HTML]{32CB00} S} & {\color[HTML]{32CB00} S} & Z                        & {\color[HTML]{3531FF} N} & \multicolumn{2}{l}{}                   & {\color[HTML]{3531FF} N} & {\color[HTML]{3531FF} N} & {\color[HTML]{FE0000} E}  \\
\multicolumn{1}{c|}{\multirow{-3}{*}{Low}}  & Mixtral        &                    & {\color[HTML]{3531FF} N} & {\color[HTML]{32CB00} S} & {\color[HTML]{32CB00} S} & {\color[HTML]{32CB00} S} & \multicolumn{2}{l}{}                   & {\color[HTML]{FE0000} E} & {\color[HTML]{32CB00} S} & {\color[HTML]{32CB00} S} & {\color[HTML]{FE0000} E} & \multicolumn{2}{l}{}                   & {\color[HTML]{FE0000} E} & {\color[HTML]{FE0000} E} & {\color[HTML]{FE0000} E} & {\color[HTML]{FE0000} E} & \multicolumn{2}{l}{}                   & {\color[HTML]{3531FF} N} & {\color[HTML]{32CB00} S} & {\color[HTML]{FE0000} E}  \\ \hline

\multicolumn{1}{c|}{High}                                       & Bloomz         & \multirow{-9}{*}{} & {\color[HTML]{32CB00} S} & {\color[HTML]{32CB00} S} & {\color[HTML]{32CB00} S} & {\color[HTML]{32CB00} S} & \multicolumn{2}{l}{\multirow{-8}{*}{}} & {\color[HTML]{32CB00} S} & {\color[HTML]{32CB00} S} & {\color[HTML]{32CB00} S} & {\color[HTML]{32CB00} S} & \multicolumn{2}{l}{\multirow{-8}{*}{}} & {\color[HTML]{FE0000} E} & {\color[HTML]{FE0000} E} & {\color[HTML]{FE0000} E} & {\color[HTML]{FE0000} E} & \multicolumn{2}{l}{\multirow{-8}{*}{}} & {\color[HTML]{FE0000} E} & {\color[HTML]{FE0000} E} & {\color[HTML]{FE0000} E}  \\  \cline{4-7} \cline{10-13} \cline{16-19} \cline{22-24} 
\multicolumn{1}{c|}{Low}                                       & Bloomz         & \multirow{-9}{*}{} & {\color[HTML]{32CB00} S} & {\color[HTML]{32CB00} S} & {\color[HTML]{32CB00} S} & {\color[HTML]{32CB00} S} & \multicolumn{2}{l}{\multirow{-8}{*}{}} & {\color[HTML]{32CB00} S} & {\color[HTML]{32CB00} S} & {\color[HTML]{32CB00} S} & {\color[HTML]{32CB00} S} & \multicolumn{2}{l}{\multirow{-8}{*}{}} & {\color[HTML]{FE0000} E} & {\color[HTML]{FE0000} E} & {\color[HTML]{FE0000} E} & {\color[HTML]{FE0000} E} & \multicolumn{2}{l}{\multirow{-8}{*}{}} & {\color[HTML]{FE0000} E} & {\color[HTML]{FE0000} E} & {\color[HTML]{FE0000} E}  \\ \hline
\end{tabular}%
}}
\caption{The Top-performing configurations based on Apriori-based Association Rules for High/Low resource level. Comparing Standard LLMs (GPT/Gemini/Mixtral) and Multilingual LLM. $confidence > 0.8, support > 0.15$. \textcolor{green}{S} / \textcolor{red}{E} - source/English language, \textcolor{black}{Z} - zero-shot (no examples), \textcolor{blue}{N} - neutral (same performance for English/Source).}
\label{tab:rules}
\vspace{-2pt}
\end{table}

\subsection {Results}
\label{chap3_results}

\renewcommand{\arraystretch}{1.1}

In Section \ref{chap_4_compare_methods}, we present the results of  \emph{selective pre-translation}  demonstrating their advantage over both \emph{direct inference} (source language only) and \emph{pre-translation} (English only). Subsequently, in Section \ref{chap_4_optimal_configuration}, we identify the optimal configurations for each task and analyze the impact of each component on the overall performance, emphasizing key considerations for effective prompting. We start off with GPT-3.5-Turbo and proceed to verify that our results generalize   to other models.

\subsubsection{Selective Pre-Translation Advantage}
\label{chap_4_compare_methods}

% Here, we display the overall advantage of selective pre-translation strategies without specifying the exact configuration, while in the next subsection, we illustrate the "optimal" configurations in different multilingual scenarios. 
Table \ref{tab:summary_results} shows each language's highest-performing configuration score among all 24 distinct configurations.\footnote{Appendix \ref{appendix:detailed_results} displays the full-fledged table of results.} Additionally, we display the improvement (\%) over \emph{direct inference} and \emph{pre-translation} for each language. 
The results indicate that 92\% of the tested languages show an improvement over the basic \emph{pre-translation} configuration. Particularly for low-resource languages like Malayalam and Telugu, the gains with \emph{selective pre-translation} are substantial, exceeding 200\% in relative improvement. Overall, when comparing selective pre-translation to  \emph{complete pre-translation}, the average improvement in low-resource languages is  65\% greater than  the average improvement in high-resource languages. 

%\paragraph{Improvement Over Direct-Inference} 

The results further reveal that 90\% of the languages show improvement over   basic \emph{direct inference}. Similar to the pre-translation approach, low-resource languages like Telugu and Assamese demonstrate a relative improvement of over 100\%. High-resource languages also show impressive improvement, albeit smaller, e.g., French and Portuguese show an improvement of over 20\% in NER.

% Figure \ref{fig:Improvement_over_monolingual} displays the percentage improvement of choosing the highest configuration score compared to the \emph{direct inference} score, highlighting that languages with lower representation achieve better improvement than those with higher. 
Overall, the table shows that \emph{selective pre-translation} can outperform both \emph{pre-translation} and \emph{direct inference}, particularly for languages considered low-resource during pre-training.

\subsubsection{The Holy Grail of  Optimal Configuration }
\label{chap_4_optimal_configuration}
Having established the advantage of \emph{selective pre-translation} in general, we now study the effects of component language selection on model performance and provide general guidelines for multilingual scenarios.\footnote{For the instruction component language, except for a slight preference for English as demonstrated in Table \ref{tab:correlation}, we did not observe a strong affinity for any language selection.}
% . Based on these observations, we narrate general rules for various multilingual scenarios. 
Table \ref{tab:correlation} shows the Point-biserial correlation between individual component selection and model performance for all the 24 configurations per language/task.\footnote{We calculate the correlation between performance and a binary vector indicating whether the component is in English.} Table \ref{tab:rules} presents the top-performing configurations, based on the highest-scoring apriori rules for multiple component selections, henceforth {\em optimal configurations}.

% See Appendix \ref{appendix:detailed_results} for
% supplementary results for other models.\reut{move to model impact}

\paragraph{Context Language Impact}

% Table \ref{tab:correlation} indicates that extractive tasks, such as QA and NER, benefit from including source-language context. This effect is particularly pronounced for low-resource languages (Class C/D), exhibiting a 70\% higher correlation coefficient with source-language context compared to high-resource languages. Conversely, tasks like abstractive summarization and NLI seem to be agnostic to context translation. Our rule-association analysis in Table \ref{tab:rules} further highlights the model's preference for source-language context in extractive tasks (NER, QA). Therefore, prioritizing source-language context for extractive tasks is generally recommended, especially for low-resource languages.

%\reut{is this only for a single model or does it generalize / change across models?} 

Table \ref{tab:correlation} shows that source language selection correlates most with model performance score in extractive tasks, such as QA (average of 0.33) and NER (average of 0.32), particularly for low-resource languages (Class C/D), which demonstrate a 70\% higher correlation coefficient compared to high-resource languages. In contrast, in tasks like abstractive summarization and NLI, we found no correlation (average of 0.05) to context language selection. Our rule-association analysis in Table \ref{tab:rules} further underscores the importance of source-language context in extractive tasks,  especially with low-resource languages.
% Therefore, it is recommended to prioritize source-language context for extractive tasks, especially for low-resource languages.

% Interestingly, the cross-task association rule that also requires source language context. 
% However, this rule has a low confidence score of 0.3, implying it's not always the case for all configurations.

\begin{figure}[t]
    \centering
    \subfloat[QA\label{fig:syntactic_heatmaps_1}]{{\includegraphics[width=0.23\textwidth,height=0.15\textwidth]{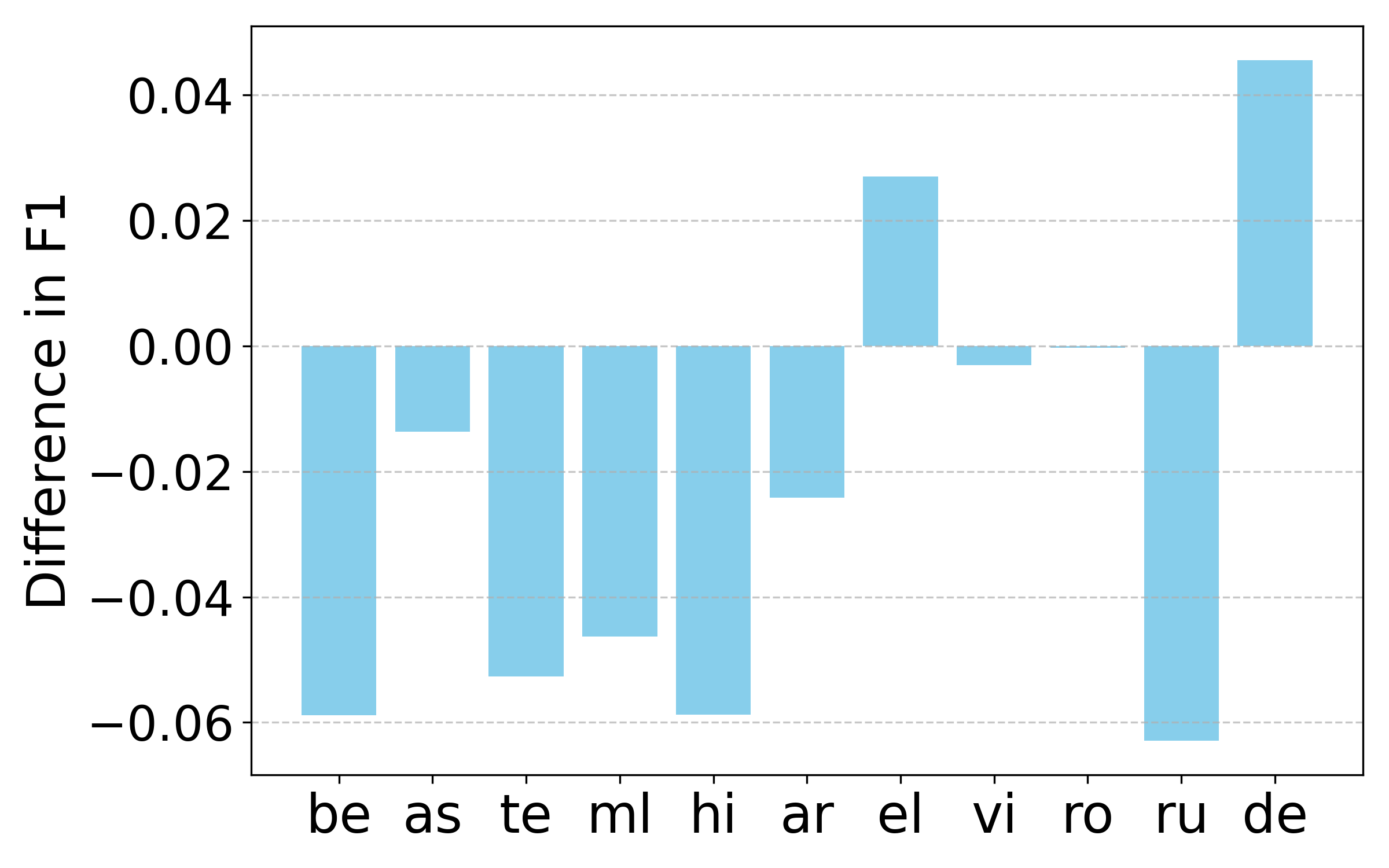} }}%
    \hfill
    \subfloat[NER\label{fig:family_langs_1}]{{\includegraphics[width=0.23\textwidth, height=0.15\textwidth]{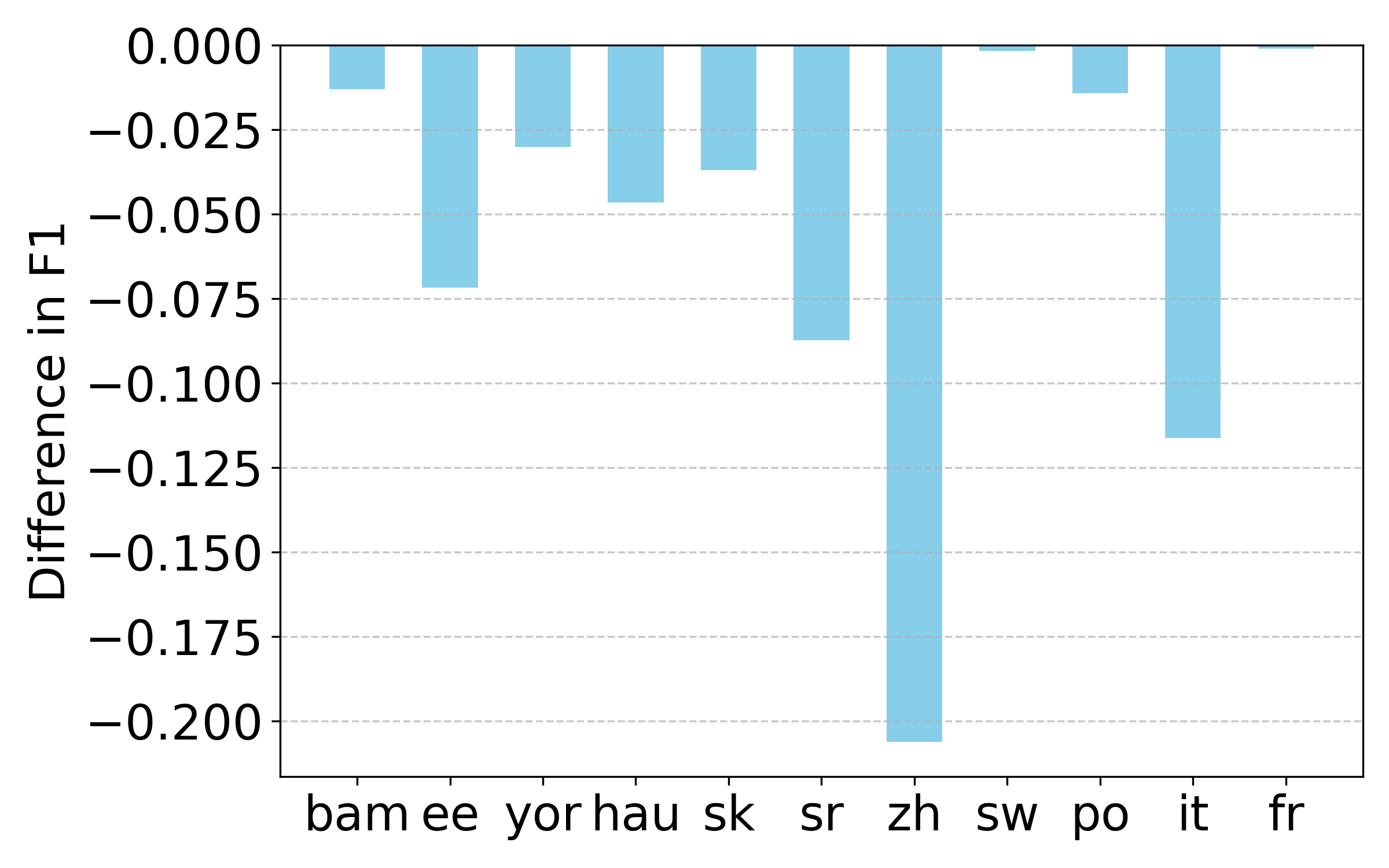} }}%
    \hfill
    \subfloat[NLI\label{fig:syntactic_heatmaps_2}]{{\includegraphics[width=0.23\textwidth,height=0.15\textwidth]{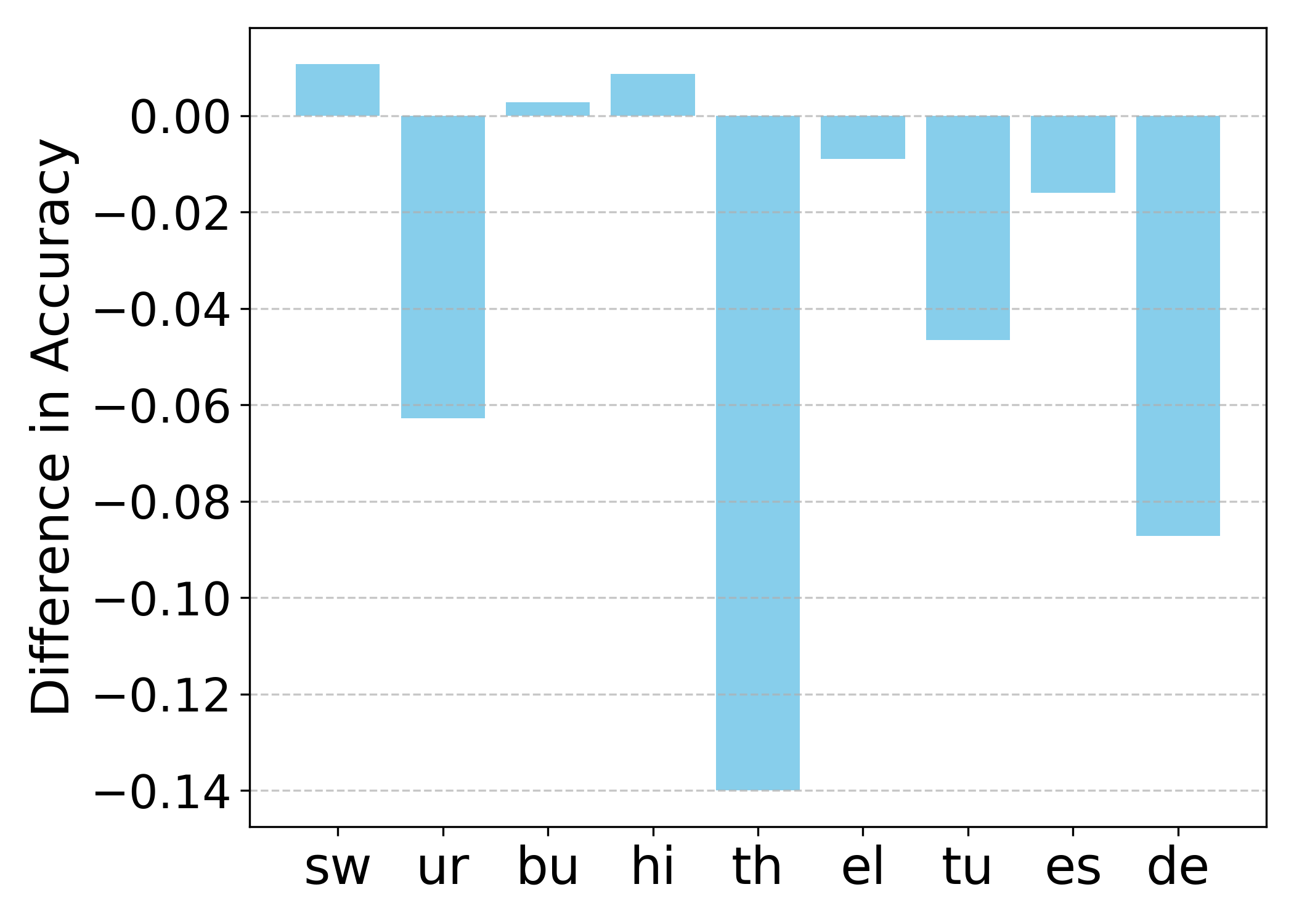} }}%
    \hfill
    \subfloat[Summarization.\label{fig:family_langs_2}]{{\includegraphics[width=0.23\textwidth, height=0.15\textwidth]{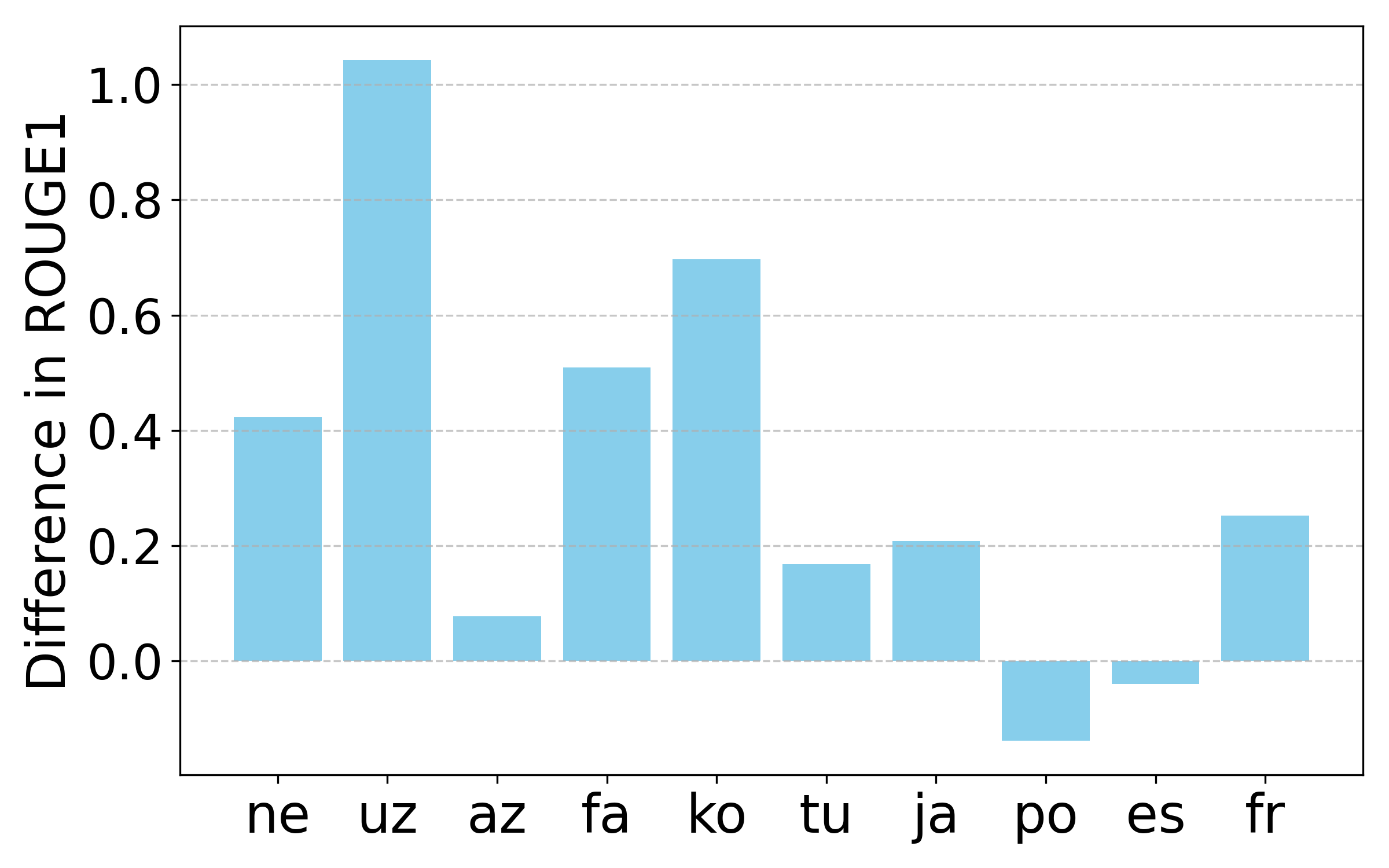} }}%
    \caption{Performance Gap Analysis for the Examples language (English minus Source). Left to right X-axes order indicates Low to High Resource Level. Y-axes indicate language preference: positive values for the source language, and negative for the English language.}%
    \label{fig:english_shot_target_shots}%
    \vspace{-2pt}
\end{figure}

% \paragraph{Examples Language Impact} Analysis of rule associations in Table \ref{tab:rules} reveals that extractive tasks (NER, QA) benefit from incorporating source-language examples. We hypothesize that because NER is a region-specific task that requires the model's pre-trained knowledge, examples in the source language help the model augment its knowledge base in the specific language. Figure \ref{fig:english_shot_target_shots} aligns with these findings, depicting a similar performance gap between few-shot English examples and source-language examples.  

\paragraph{Examples Impact}

In general, the top-performing configurations of 
the GPT model in Table \ref{tab:rules}
%\reut{I dont understand how to read this table as "rules"}
%\reut{plus we have lots of ablation for the "english centric" and extremely little breakdown on multilingual - why?}
show that the optimal configurations are those that include examples, i.e., a few-shot rather than zero-shot setup (Appendix \ref{appendix:optimal_configuration} further supports incorporating examples in prompts, especially for high-resource languages). 
%\paragraph{Examples Language Impact} 
Concretely concerning the language selected for the examples, the optimal configurations in Table \ref{tab:rules} show that extractive tasks as NER perform better with source-language examples, possibly due to NER's dependence on region-specific or culturally-relevant knowledge. Also, the performance gap analysis in Figure \ref{fig:english_shot_target_shots},  shows that, for extractive tasks, prompts with examples in the source language perform better than those with English examples, especially for low-resource languages (See \ref{fig:english_shot_target_shots}(a)/\ref{fig:english_shot_target_shots}(b)).

\paragraph{Output Language Impact}
Unlike context and examples, the output depends on the model generations's grammaticality and fluency. The best-performing configurations for the GPT model in Table \ref{tab:rules} indicate that for extractive tasks, source-language output is beneficial across all languages. Interestingly, despite context mismatches, NER in low-resource languages also benefits from English output. For generative tasks such as summarization, model output in English 
%\reut{do you mean you assess the output in english? I thought you translate back to source. what is the rouge measuring exactly?} 
performs better due to the model’s stronger capabilities in English,  {\em even though} we back-translate the output to source prior to evaluation. Thus,  while it is fine  in such generative tasks to instruct the model to generate outputs in the source language for high-resource languages, it appears better to generate in English in the low-resource case.

% However, Appendix \ref{appendix:output_english_output_target_appendix} displays the performance gap and shows that output for NER is more ambiguous. 

\subsection{Beyond Configuration: Key Factors}

Having analyzed the impact of the components' language selection, we discuss key additional factors influencing the efficacy of our approach.\footnote{See also Appendix \ref{appendix:script_impact} for script impact.}

\paragraph{Pre-Training Data Size Impact}
% Table \ref{tab:summary_results} presents the highest-prompting prompt configuration for each language and task and shows that for QA, summarization and NER the general trend is that classes A and B (High-Medium resource) achieve better results than those in classes C and D (Low resource), indicating that more pre-trained data yields better performance. However, for NER we found not notable exceptions exist, i.e., Hausa and Ewe (Class C/D) achieve better results on the NER task compared to Swedish and Chinese (Class A/B). For NLI task there is no clear train where several class C/D languages outperform A/B in NLI. While pre-trained data size matters, selective pre-translation can help low-resource languages surpass high-resource ones in specific tasks.

Table \ref{tab:summary_results} presents the optimal prompt configuration scores per language and task. For QA, summarization, and NER, the general trend indicates that even for the optimal pre-translation configuration,  classes A and B (High-Medium resource) achieve better results than classes C and D (Low resource),
%\reut{on what configurations are these types better?} 
%suggesting that larger amounts of pre-trained data lead to improved performance. 
However, a few exceptions exist, i.e., in Hausa and Ewe (Class C/D) we see better results on the NER task compared to Swedish and Chinese (Class A/B). For the NLI task we found no trend where a  class C/D languages outperform A/B languages. So, while pre-trained data distribution matters, selective pre-translation can help low-resource languages match the results of higher-resource ones in specific tasks.

\begin{figure*}[th]
\scalebox{1.0}{
\centering
\includegraphics[width=1\textwidth]{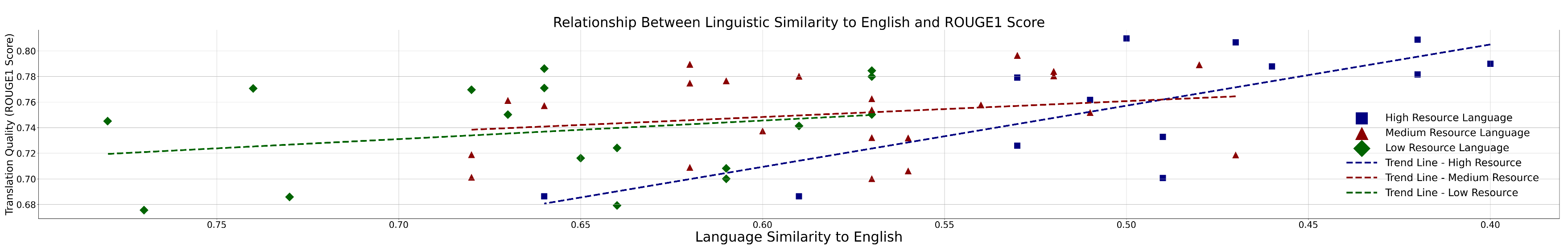}
}
% \vspace*{-1mm}
%\footnotesize{
\caption{Scatter plot showing the relationship between syntactic similarity to English (further right is more similar) and translation quality (ROUGE) for four language resource subsets (represented as distinct four colored shapes).  Each  dot represents a different language. Positive linear regression shows an upward trend. }\label{fig:syntactic_similarity_translation}
%}
\end{figure*}

\paragraph{Linguistic Similarity to English Impact} We used pre-computed syntactic similarities to English from the URIEL dataset \citep{littell2017uriel} and calculated the Pearson correlation for each task between the best-performing configuration scores (top score per language) and the syntactic similarity of these languages to English. A moderate correlation (0.42) for summarization shows that syntactic similarity to English positively correlates with performance. NER also shows moderate correlations, suggesting models better identify entities when texts share syntactic features with English. Appendix \ref{appendix:factors_explaning_performance} further details these results.

%\reut{I didnt find there the details I am looking for: eg which features and what is the calculation}
%\reut{figure 8 tables are too tiny. also other tables can benefit from larger font}

\paragraph{Standard Model Impact}

In the previous section, we assessed the selective pre-translation strategies using the GPT model. In this section we check whether  these strategies generalize to other  LLMs. Table \ref{tab:rules} displays the optimal configurations per task and language for Gemini and Mixtral. We  see that the {\em preference for source language in extractive tasks} (for context, examples, and output)  holds across all three models. Additionally, outputting in English while keeping the {\em context in the source language for NER} in low-resource languages is consistent. In NLI, models are agnostic to instruction language. However, surprisingly, in abstractive summarization, we found no clear pattern.\footnote{See Appendix \ref{appendix:detailed_results} for
additional results.}

% See Appendix \ref{appendix:detailed_results} for
% supplementary results for other models.\reut{move to model impact}

\paragraph{Multilingual Model Impact}

In addition to the standard LLMs, we evaluated BLOOMZ-7b1-mt, known for its multilingual capabilities \citep{muennighoff2022crosslingual}. Table \ref{tab:rules} displays the optimal configurations for BLOOM across all tasks. We found no distinction between resource types for this model. As shown, the preference for source language in QA is relevant here as well. Interestingly, NER can be answered in the source language, highlighting its multilingual strength. However, for generative tasks and NLI, this multilingualism diminishes, as the model tends to favor English prompts. Overall, the top-performing configurations for BLOOMZ indicate that it performs better with single-language prompts rather than with selective pre-translation prompts.

\begin{figure}[t]
  % \vspace{-1.2em} % Adjust this value to reduce the space
  \centering
\includegraphics[width=0.497\textwidth]{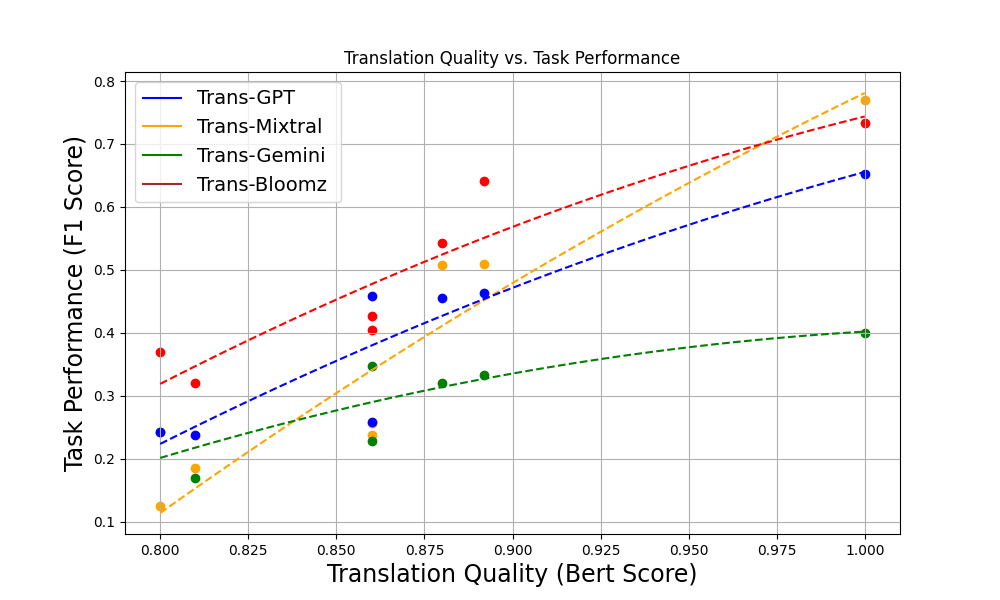}
  % \vspace{-2em} % Adjust this value to reduce the space
  \caption{Correlation between translation quality (BERTScore) and accuracy (F1) for Pre-Translation-Zero-shot prompting, each  dot is a different language.}
  \label{fig:bleu_f1}
   % \vspace{-1em}
\end{figure}

% \section{Translation Quality Impact}
\section{Translation: Key to Pre-Translation}
\label{chap_5}
% The correlation with the High languages is the strongest, with a coefficient of -0.73 and a p-value of 0.004. The negative correlation indicates that languages with smaller linguistic distances (higher similarity) to English tend to achieve better machine translation to English.

In the previous Section we examined various factors affecting model performance for selective pre-translation strategies. Since translation forms the foundation of  selective pre-translation, this Section focuses on a key question: Are these factors primarily due to the limitations of LLMs, or are due to the quality of the pre-translations themselves?
% attested for that language? 
To address this, we  first isolate the impact of these factors on translation quality through a controlled experiment. Subsequently, we  investigate how translation quality, independently of other factors, influences downstream tasks in our setup.

\subsection{Experimental Setup}
% \paragraph{Goal} 

% \paragraph{Models} 

% To evaluate the translation quality we used two translator engines: \emph{Google Translate API}\footnote{We used the open-source library 
% \url{https://pypi.org/project/easygoogletranslate/}.} and \emph{Bing Translator API} over Azure. 

\paragraph{Factors Affecting Translation Quality}
We evaluated Google Translate performance  on the
%\reut{of what? which translator? and is it the one used in sec 4?} 
FLORES-200 validation set \citep{guzman-etal-2019-flores}, analyzing 91 languages across all resource levels, each with 997 sentences paired with  their  English translations. We compared machine translations to human-generated references using: (1) n-gram matching metrics -- Meteor \citep{banerjee2005meteor}, ROUGE \citep{lin2004rouge}, and BLEU \citep{papineni2002bleu}, and (2) neural network-based evaluation metrics -- BertScore \citep{zhang2019bertscore} and Comet \citep{rei2020comet}. Additional experiments using other translation models are in Appendix \ref{appendix:mt_engines_comparsion}.

\paragraph{Impact on Downstream tasks}
We used the QA dataset XQuAD, with 300 parallel sentences in English and other languages.
% with parallel splits in English and other languages, each containing 300 sentences.
We compared the zero-shot \emph{pre-translation} using output in the source language with the \emph{selective pre-transaltion} optimal approach using GPT-3.5-Turbo (0125).

% \paragraph{Data} For determining the quality of languages we evaluated translation quality using the FLORES-200 validation set \citep{guzman-etal-2019-flores}, focusing on 91 languages from all resouce levels, each with 997 sentences translated to English.\footnote{All the selected languages supported by Google Translate \url{https://cloud.google.com/translate/docs/}} For the second experiment, we used the XQuAD dataset for the QA task, with parallel splits in English and other languages, each containing 300 sentences. We examined the \emph{pre-translation} approach in a zero-shot setting with English output.

% \paragraph{Evaluation}
% We compared machine translations to human-generated references using: (1) n-gram matching metrics -- Meteor \citep{banerjee2005meteor}, ROUGE \citep{lin2004rouge}, and BLEU \citep{papineni2002bleu}, and (2) neural network-based evaluation metrics -- BertScore \citep{zhang2019bertscore} and Comet \citep{rei2020comet} using Google Translate. Additional experiments using other translation models are in Appendix \ref{appendix:mt_engines_comparsion}.

 % We evaluate translation quality by translating each sentence using both Google and Bing translation engines \footnote{Appendix \ref{appendix:mt_engines_comparsion} provides additional analysis and a full breakdown of the comparative finding}.

\subsection{Results}
\label{chat5_results}

\paragraph{Factors Affecting Translation Quality}

 % To determine the impact of the size of a language in the pre-trained data, we used the data ratio of the language in the GPT-3 unlabeled pre-training data. 
Our analysis focuses on the factors influencing model performance discussed in Sec.~\ref{chap_4} and we explore their impact on translation quality. We focus on two factors: \textbf{\em language resource levels (high/low)} and \textbf{\em linguistic similarity to English}. First, we found that the average quality for high-resource languages was higher compared to low-resource languages (0.75 vs 0.73), although correlation between resource level and quality wasn't significant. As for   {\em linguistic similarity to English} we used the pre-computed linguistic similarities from the URIEL dataset \citep{littell2017uriel}\footnote{\url{https://github.com/antonisa/lang2vec}} and calculated the correlation between syntactic similarity to English and the translation quality for each language. Figure \ref{fig:syntactic_similarity_translation} shows a positive correlation (coefficient = 0.33, p-value = 0.01) between syntactic similarity to English and translation quality (ROUGE-1). This correlation is particularly strong for high-resource languages (coefficient=0.73, p-value=0.004). Additional correlation results are detailed in Appendix \ref{appendix:linguistic_similarity_to_english_correlation}.

\paragraph{Impact On Downstream Tasks}

To isolate translation quality as the sole direct factor influencing model performance, we translated the entire prompt into English. Since the original input differed only in language, not content, any variation in the processed input can be attributed solely to the quality of the translation. This approach %eliminates the direct impact of factors such as language resource size and similarity to English, allowing 
allows us to directly measure the correlation between translation quality and model performance across various tasks. 
Figure \ref{fig:bleu_f1} shows the correlation between translation quality (BERTScore) and model performance (accuracy) for each language. Our results show that {\em higher translation quality goes hand in hand with improved task performance}. The overall Pearson correlation is 0.233 (p < 0.001). 
However, when assessing the  same tasks with  \emph{selective pre-translation} 
instead of a \emph{completely pre-translated} prompt, we found a low correlation of 0.05 (p< 0.05) between the translation quality and  task performance, while  selective pre-translation outperforms the fully translated prompt. 
This disparity shows that the \emph{selective pre-translation} method effectively neutralizes translation issues.  By strategically choosing which prompt components to translate, we can make pre-translation useful for languages with lower translation quality.

In sum, our findings demonstrate that factors influencing downstream tasks, such as high resource level and  similarity to English, are positively correlated with translation quality. We further show that  \emph{selective pre-translation} can mitigate the negative effects of poor translation quality. These two findings underscore the importance of investing in high-quality translation, and on the other hand, prioritizing the \emph{selective pre-translation} approach in languages where machine translation is sub-optimal.

\section{Related Work}

%\subsection{The Rise of Multilingual LLMs}

With over 7,000 languages spoken globally \citep{anderson2010many},   the growing use of diverse languages have fueled the demand for multilingual LLMs. Progress in this field stems from two primary efforts: (1) developing dedicated monolingual models for low-to-medium-resource languages \citep{seker2022alephbert, cui2023efficient, andersland2024amharic}, and (2) creating multilingual LLMs with pre-trained data encompassing multiple languages  \citep{qin2024multilingual, jiang2024mixtral}. 

The ability of Multilingual LLMs to operate in % and transfer knowledge between,
different languages  \citep{raffel2020exploring, conneau2019unsupervised, chowdhery2023palm} comes from two sources: (1)  training or fine-tuning on multilingual data in order to achieve multilingual proficiency
\citep{xue2020mt5, chen2021zero, le2023bloom, shaham2024multilingual, muennighoff2022crosslingual}, and (2) utilizing prompting techniques to harness the model's inherent multilingual capabilities without modifying parameters during inference \citep{brown2020language}. This latter approach has gained popularity due to its efficiency and applicability to a wider range of use cases.

%Following these developments, benchmarks for evaluating LLMs have been proposed to measure cross-lingual transfer, including for low-resource languages \citep{hu2020xtreme, liang2020xglue} and benchmarks focusing on specific language families such as Indian languages  \citep{kakwani2020indicnlpsuite} and African languages \citep{ogundepo2023afriqa}. 

%\subsection{Multilingual Prompting Approaches}

For the latter, to improve the multilingual capabilities of  LLMs
researchers
%\reut{if this pertains to item (2) above, say it}
  developed various prompting methods. \citet{huang2023not} introduced XLT, a cross-lingual prompt that directs LLMs to function as experts in a specific language through a process involving problem-solving and cross-lingual thinking. \citet{zhao2021discrete} employed discrete and soft prompting techniques and showed that few-shot non-English prompts outperform finetuning in cross-lingual transfer. \citet{shi2022language} found that chain-of-thought (CoT) prompts lead to multilingual reasoning abilities in LLMs, even in under-represented languages. Another strategy is \emph{pre-transaltion} which translates the entire prompt to English \citep{chowdhery2023palm, qin2023chatgpt, ahuja2023mega}. A more nuanced approach, \emph{selective pre-translation}, translates part of the prompt into English, for instance, \citet{liu2024translation} translated only the instruction, and \citet{ahuja2023mega} translated the few shot examples. While
these use cases lack a systematic research foundation, in this study, we systematically study pre-translation configurations to provide evidence-based recommendations for optimal use.
%\reut{this background section feels redundant  and breaks the flow- it should be moved to after section 5 and be called "related work". Also, for works you mention in 2.2. in this related work section need to state how they are similar or different than this paper }

\section{Conclusion}

In this work we formalize and comprehensively assess selective pre-translation prompting strategies for LLMs in multilingual settings. With four tasks, six dataset collections, three models, and 35 languages, we deliver the first systematic evaluation, to our knowledge, of all existing prompt configurations of pre-translation. We demonstrate that \emph{selective pre-translation}  consistently outperforms both \emph{pre-translation} of the entire prompt and \emph{diret-inference} in the source language, establishing the efficacy of \emph{selective pre-translation} in both the high- and low-resource cases. Additionally, we show 
that translation quality significantly
affects  performance and that 
selective pre-translation can mitigate the negative effects of suboptimal translations.

\section*{Limitations}

\paragraph{Subset of LLMs}
This study aims to systematically assess the effectiveness of various prompting strategies across different tasks and LLMs. Due to resource limitations, it was not possible to evaluate more advanced models such as GPT-4 or GPT-4o. However, we endeavored to cover several LLMs representing different architectures. Additionally, the choice of Bloom as our multilingual model is based on previous works \cite{bawden2023investigating, nezhad2024drives}. We make our evaluation framework, code, configurations, and execution pipeline, for open public use, allowing to extend the investigation to more and newer models.

\paragraph{LLM Adherence and Impact on the Output}

In our evaluation, we attempted to influence the output by instructing the model to generate a response in a specific language. However, the model occasionally did not follow these instructions, producing output in a different language, which could impact the results. Appendix \ref{appendix:error_analysis} provides error analysis of the various issues we encountered.  

\paragraph{Evaluation}
Metrics based on n-gram matching, such as ROUGE \citep{lin2004rouge}, are commonly used for evaluating summarization quality in English. However, these metrics can be problematic when applied to morphologically rich languages (MRL) such as Persian, which have more flexible word order compared to English. Additionally, their morphological richness means that the same concept can be expressed in multiple ways due to variations in prefixes, suffixes, and root conjugations.

\paragraph{Translation Quality's Impact on Downstream tasks}

Our analysis of the impact of translation quality impact on downstream tasks in section \ref{chat5_results} was constrained by the scarcity of datasets with parallel splits for English and other languages, limiting our evaluation to the QA task.  
Future research should incorporate a wider array of datasets and tasks to validate and expand upon our findings.

\paragraph{Pretrained Data Distribution Details}

In our experiments, we evaluated four models and grouped the languages based on GPT-3's pre-training data distribution information. Ideally, we would split the languages according to each model's data distribution. However, to our knowledge, only GPT-3's pre-training data distribution is publicly shared. Explicitely testing different language  distributions is desired but resource intensive, and is left for future research.

\section*{Acknowledgements}
This research has been funded by a grant from the Israel Science Foundation (ISF) grant number 670/23 as well as a grant from the Israeli Ministry of Science and Technology (MOST), and a KAMIN grant from the Israeli Innovation Authority, for which we are grateful. We are further grateful for a generous VATAT grant to the BIU NLP team which contributed resources for computation, annotation  and human evaluation in this project.

% \section*{Acknowledgements}

\bibliography{anthology}
\bibliographystyle{acl_natbib}

\appendix
\label{appendix:data_collection}

\begin{figure*}[ht]
    \centering
    \subfloat[<English,English,English,English>\label{fig:syntactic_heatmaps_1}]{{\includegraphics[width=0.32\textwidth]{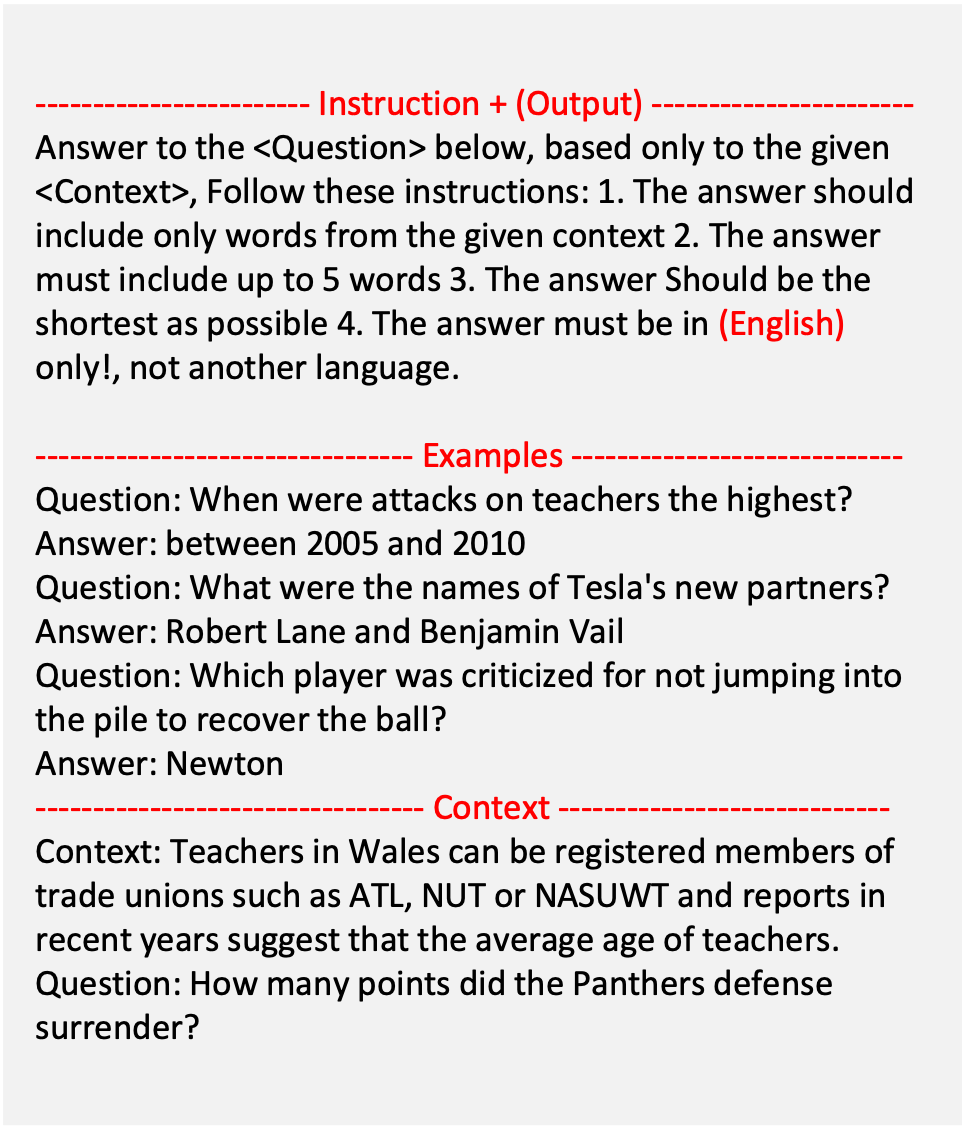} }}%
    \hfill
    \subfloat[<German,English,German,English>\label{fig:family_langs_1}]{{\includegraphics[width=0.32\textwidth]{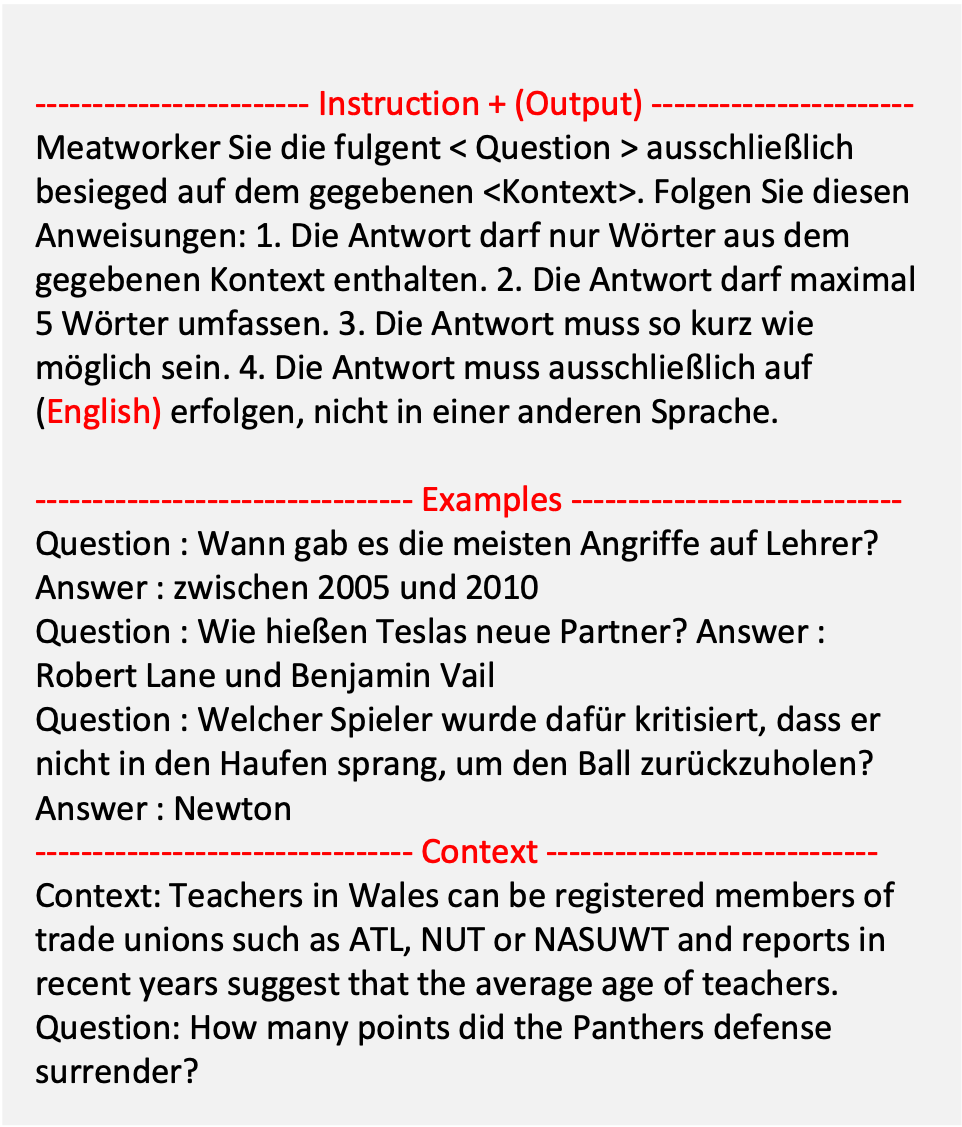} }}%
    \hfill
    \subfloat[<German,German,English,English>\label{fig:syntactic_heatmaps_2}]{{\includegraphics[width=0.32\textwidth]{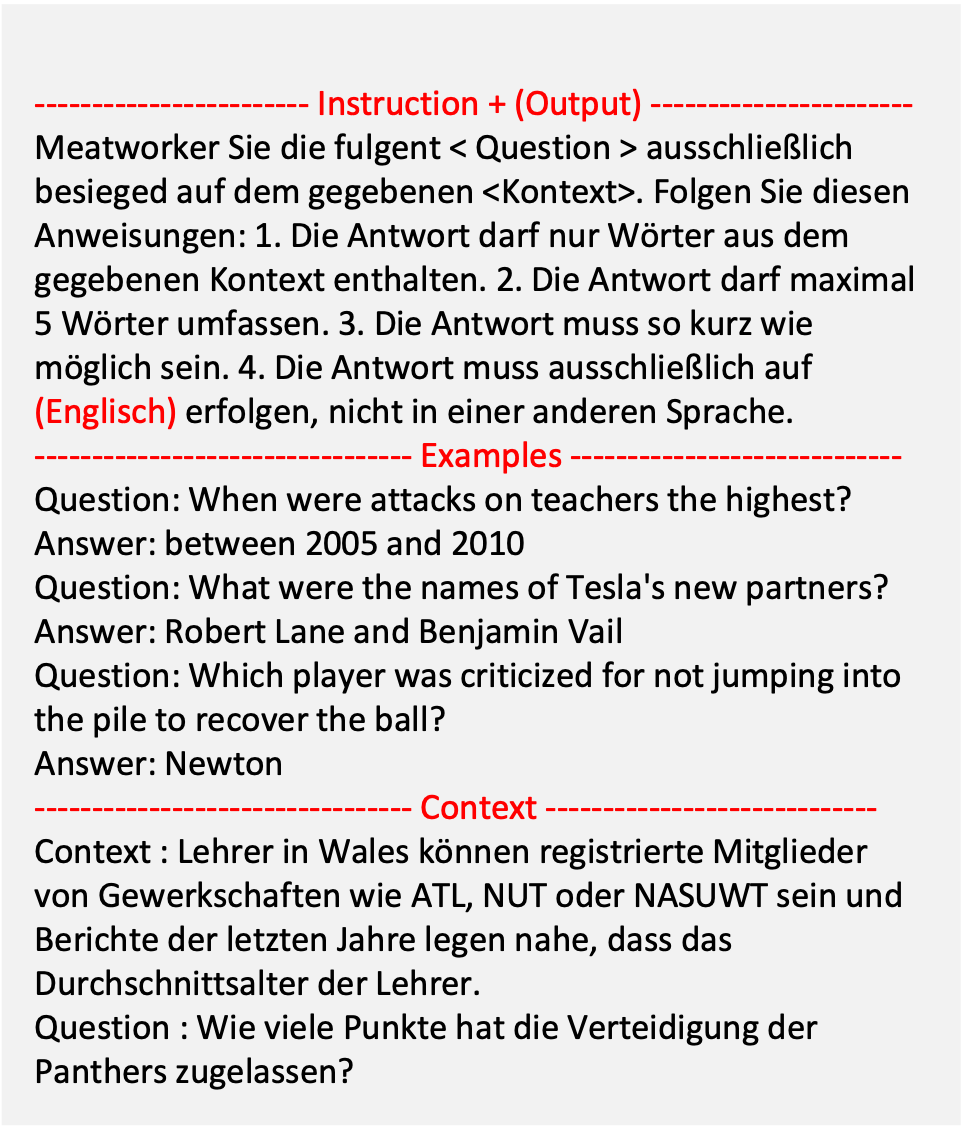} }}%
    \caption{Examples of 3 configurations of German. Each configuration is in the following format {\em <Instruction,Context,Examples,Output>}}.%
    \label{fig:prompting_format}%
\end{figure*}

% \begin{figure*}[t]
%     \centering
%     \includegraphics[width=1.0\linewidth, height=0.1\textheight]{improvement_over_english.png}
%     \caption{Percentage of improvement over \emph{pre-translation} configuration, when using the highest configuration for each languages/task For GPT-3.5-Turbo. The bars are color-coded based on the norm of the log of the number of tokens in the pre-trained data of GPT-3, as elaborated in Table \ref{tab:languages_classes}}
%     \label{fig:improvement_over_english}
% \end{figure*}

\begin{table}[]
\centering
\resizebox{\columnwidth}{!}{%
\begin{tabular}{llll}
\multicolumn{4}{c}{\textbf{All Configurations}}                                                                                                              \\
\multicolumn{1}{c}{\textit{Instruction}} & \multicolumn{1}{c}{\textit{Context}} & \multicolumn{1}{c}{\textit{Examples}} & \multicolumn{1}{c}{\textit{Output}} \\ \hline
Source                                   & Source                               & -                                    & English                             \\
English                                  & Source                               & -                                    & English                             \\
Source                                   & Source                               & Source                               & English                             \\
English                                  & English                              & English                              & Source                              \\
Source                                   & English                              & Source                               & English                             \\
English                                  & Source                               & Source                               & Source                              \\
English                                  & Source                               & -                                    & Source                              \\
Source                                   & Source                               & Source                               & Source                              \\
English                                  & Source                               & English                              & Source                              \\
Source                                   & Source                               & -                                    & Source                              \\
English                                  & English                              & Source                               & Source                              \\
English                                  & Source                               & English                              & English                             \\
English                                  & English                              & -                                    & English                             \\
Source                                   & English                              & -                                    & English                             \\
Source                                   & English                              & -                                    & Source                              \\
Source                                   & Source                               & English                              & English                             \\
English                                  & English                              & -                                    & Source                              \\
English                                  & Source                               & Source                               & English                             \\
Source                                   & Source                               & English                              & Source                              \\
Source                                   & English                              & Source                               & Source                              \\
English                                  & English                              & English                              & English                             \\
Source                                   & English                              & English                              & English                             \\
English                                  & English                              & Source                               & English                             \\
Source                                   & English                              & English                              & Source                              \\ \hline
\end{tabular}}
\caption{All Valid Configurations (24)}
\label{tab:all_configurations}
\end{table}

\section{Selective Pre-Translation Evaluation}

\subsection{Experimental Setup}
\label{appendix:modular_prompting_experimental_setup}

\paragraph{Models}
\label{appendix:models_platforms} 
To query GPT-3.5-turbo (0125), we used the Azure platform via the API\footnote{\url{https://learn.microsoft.com/en-us/azure/ai-services/openai/concepts/models}}. For Mixtral-8x7B-287 Instruct-v0.1, we utilized the API platform provided by Together.ai\footnote{\url{https://www.together.ai/}}. For Gemini-1.0-pro, we accessed the API through Google AI Studio\footnote{\url{https://aistudio.google.com}}. Lastly, for bloomz-7b1-mt, we used deployed the model on Hugging Face\footnote{\url{https://huggingface.co/bigscience/bloomz-7b1-mt}}

\paragraph{Prompt Creation}
For constructing the prompts we used the LangChain library\footnote{\url{https://pypi.org/project/langchain/}} which enables us to build and validate prompts dynamically for both zero-shot and few-shot templates. For creating the instructions, we initially used ChatGPT to generate them and then fine-tuned them based on quality analysis from our experiments.

\paragraph{Normalization And Formatting}
Before evaluation, we normalized the model's output, with each task following a unique normalization process. For the QA task, for example, we converted the text to lowercase and removed punctuation, articles, and extra whitespace. In the Summarization task, we removed prefixes like 'The Summary:'. For the NER task, we converted the model's output into a list of tuples, each in the format (Tag, Entity). After normalization, additional formatting was applied if necessary. For instance, in the NER task, we transformed the normalized output into a list in the BIOSE format, identifying the entities in the original sentence and converting each entity prediction to its correct format based on its position (e.g., B-ORG for the first entity tagged as 'ORG').
\label{appendix: normalization_and_formatting}

\subsection{Configuration Format}

We define a specific {\em selective pre-translation configuration} as \( C_{i} = \langle \text{I}^l, \text{X}^l, E^{l}_{n}, O^l \rangle \), \( n \geq 0 \), \( l \in \{ e, s \} \). Each configuration contains 4 components: instruction, context, examples, and output. Figure \ref{fig:prompting_format} displays examples for 3 configurations in the German language. See Table \ref{tab:all_configurations} for a list of all the configurations.
\label{appendix:configuration_format}

\paragraph{Python Libraries In Use}
For evaluation of the different models, we used the most common ROUGE package for non-English papers\footnote{\url{https://github.com/csebuetnlp/xl-sum/tree/master/multilingual_rouge_scoring}}. for loading and processing the data, we used NumPy\footnote{\url{https://pypi.org/project/numpy/}}
For help with writing the code, we used assistance from ChatGPT.

% \paragraph{Pre-Trained Data Class Categorization} 

% Table \ref{tab:languages_classes} includes our categorization of languages into four classes, differentiated by their data ratios in the GPT-3 pre-trained data. We chose the pre-trained data of GPT-3 due to its extensive coverage and diverse language representation, enabling us to categorize the languages into classes based on their data ratios. To determine the subsets, we utilized the class division proposed by \citet{lai2023chatgpt} and modified class D to include only the languages that are unrepresented in the GPT-3 pre-trained data. To our knowledge, we are the first to specifically reference this subset of languages. Notably, other divisions exist; for example, \citet{joshi2020state} proposed dividing languages into classes based on the number of speakers. However, this approach does not faithfully represent language diversity in LLMs, which are more influenced by the language's data availability than by how well it is spoken among people.
% \label{appendix:languages_categorization}

\subsection{Analysis Methods}

\subsubsection{Rule Association And Apriori Algorithm}
\label{appendix:rule_association_recap_implementations}
Association rule mining, one of the most important and well-researched techniques of data mining, was first introduced by \citet{agrawal1993mining}. It aims to extract interesting correlations, frequent patterns, associations, or casual
structures among sets of items in the transaction databases or other data repositories. 

\paragraph{Apriori algorithm}
The Apriori algorithm is a popular approach for mining association rules. It works by identifying frequent itemsets, which are groups of items that appear together in a dataset with a frequency above a specified threshold. The algorithm then generates association rules from these frequent itemsets, highlighting the likelihood of one item being present given the presence of another item. Apriori uses a bottom-up approach, gradually building larger itemsets from smaller ones while pruning those that do not meet the minimum support threshold.

In our analysis, we reported the following measures: 
(i) \textbf{Support}: $s(X) = \frac{\sigma(X)}{N}$, where $\sigma(X)$ is the number of transactions in which X appears  and $N$ is the total number of transactions.

(ii) \textbf{Confidence}: $c(X \rightarrow Y) = \frac{\sigma(X \cup Y)}{\sigma(X)}$, measures the probability of occurrence of itemset $Y$ with itemset $X$.

\label{tab:apriori_technical_into}

\paragraph{Implementation Details}

To implement the Rule Association algorithm, we created a DataFrame for each task's results using pandas DataFrames\footnote{\url{https://pypi.org/project/pandas/}}. Each DataFrame contains the results for all the configurations for every language. Subsequently, we binned each score column into three bins - high, medium, and low, based on the 30th and 60th percentiles. Later, we merged all the data frames based on the configuration name. Then we used the apriori algorithm from the efficient-apriori\footnote{\url{https://pypi.org/project/efficient-apriori/}} library, which produces two outputs - itemsets and rules. Later, we filtered weak rules (support > 0.05 \& confidence > 0.75).
% \label{tab:apriori_appendix}
\label{appendix:rule_association}

\begin{figure*}[]
    \centering
    \subfloat[QA\label{fig:syntactic_heatmaps_1}]{{\includegraphics[width=0.23\textwidth]{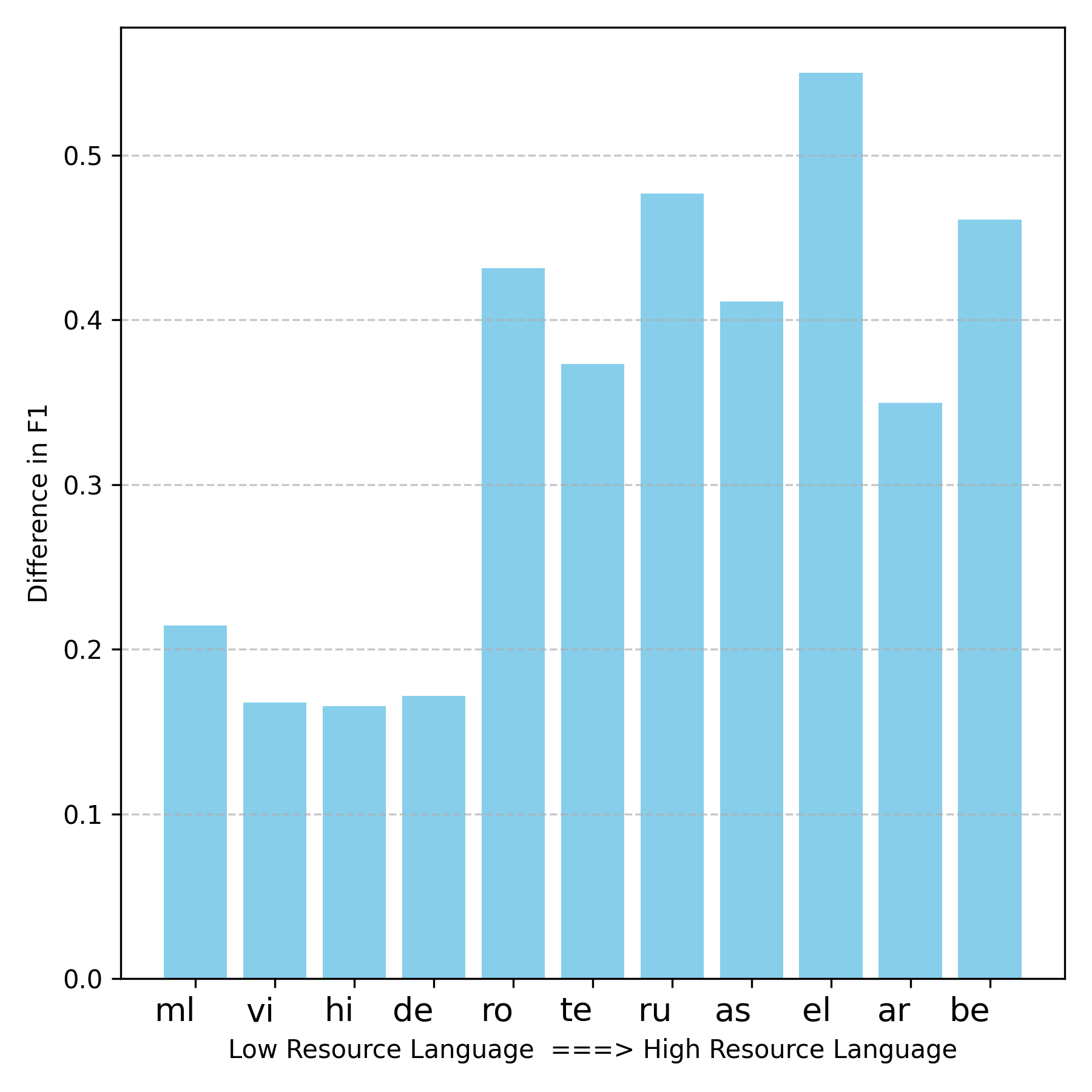} }}%
    \hfill
    \subfloat[NER\label{fig:family_langs_1}]{{\includegraphics[width=0.23\textwidth]{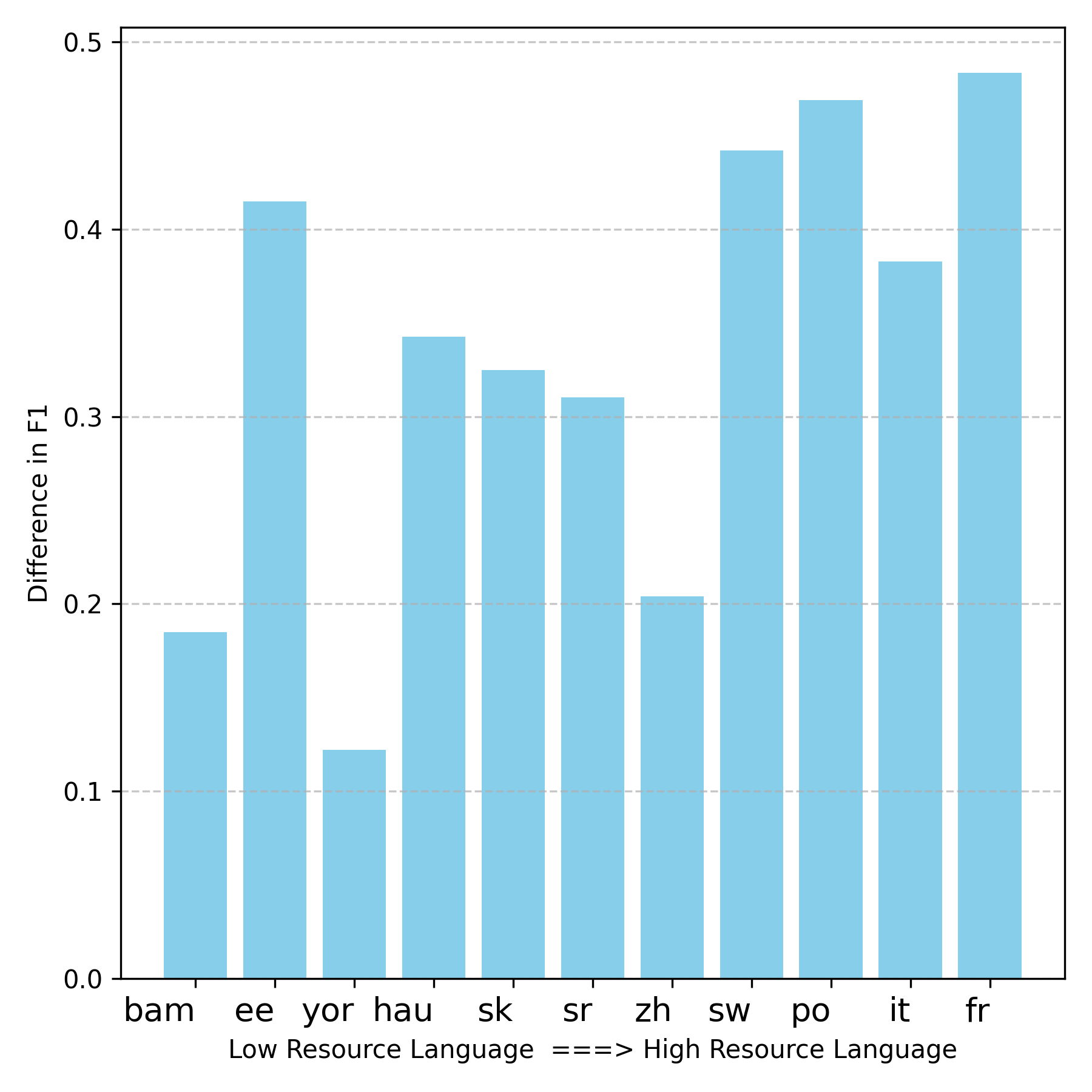} }}%
    \hfill
    \subfloat[NLI\label{fig:syntactic_heatmaps_2}]{{\includegraphics[width=0.23\textwidth]{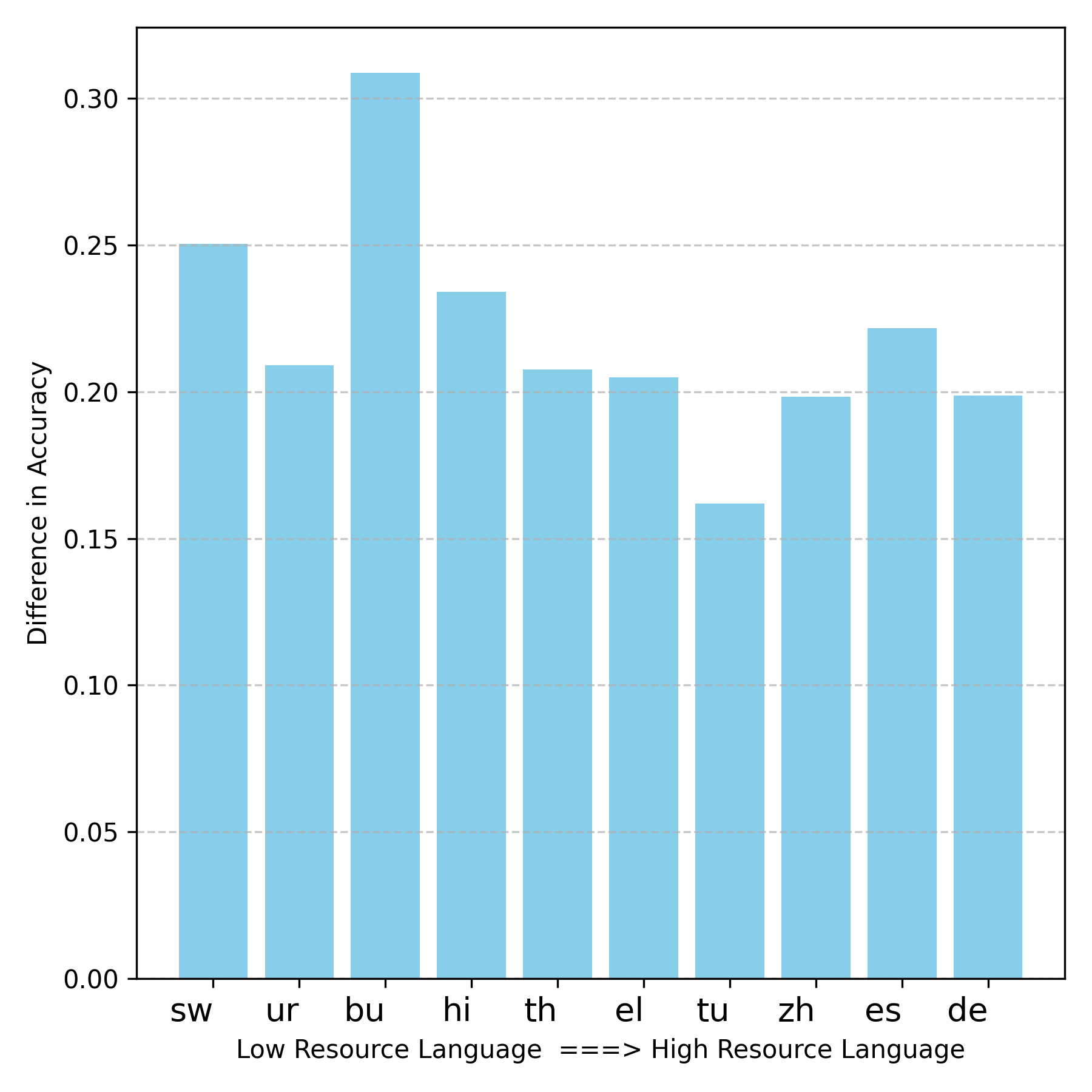} }}%
    \hfill
    \subfloat[Summarization\label{fig:family_langs_2}]{{\includegraphics[width=0.23\textwidth]{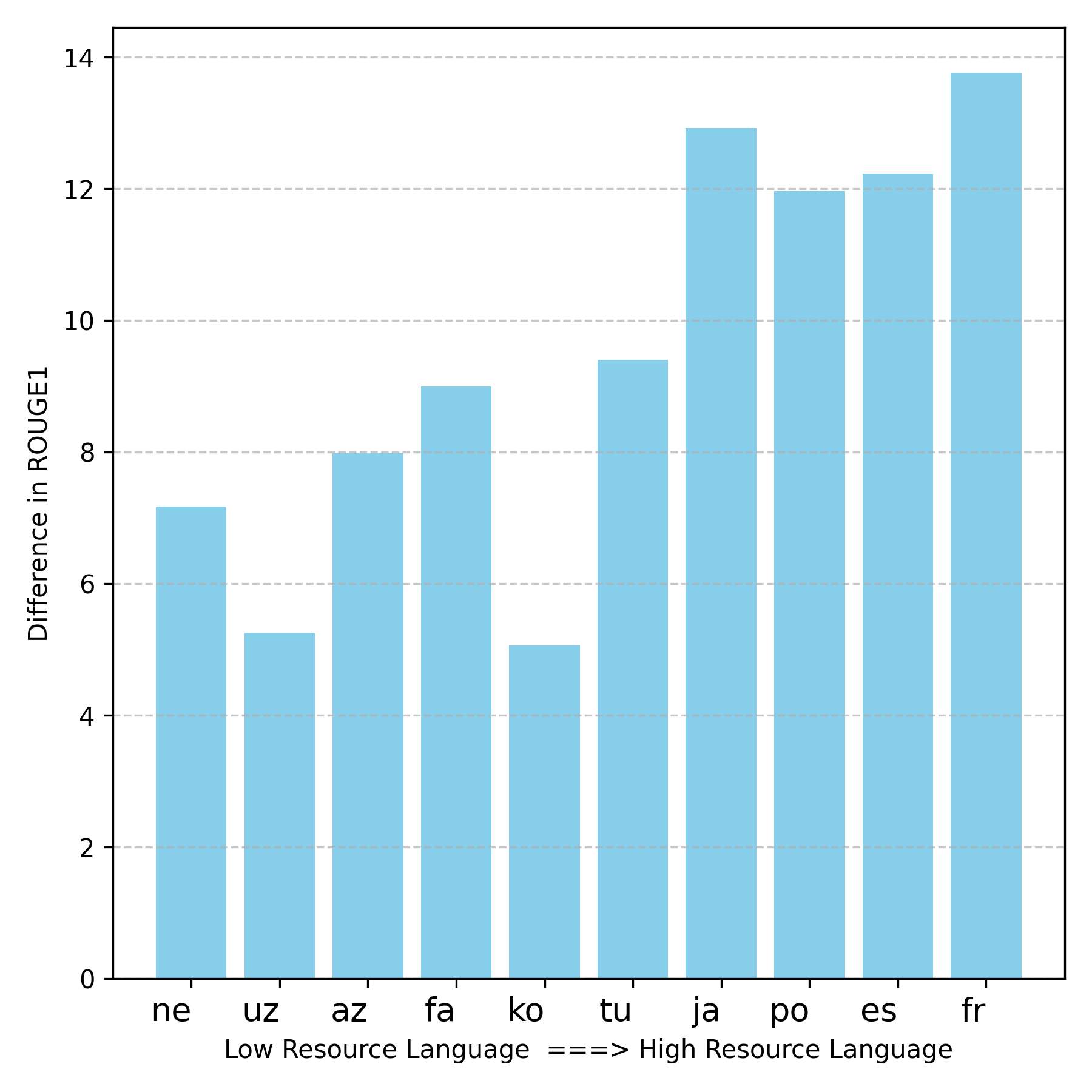} }}%
    \caption{Few-Shot and Zero-Shot Performance Gap (Few-Shot - Zero-Shot) for each task/language.}%
    \label{fig:zero_shot_few_shot}%
\end{figure*}

% \begin{table}[!ht]
% \centering
% \scalebox{0.9}{
% \begin{tabular}{lccc}
% \Xhline{6\arrayrulewidth}
% \textbf{Model}        & \textbf{QA} & \textbf{NER} & \textbf{Summarizatin} \\ \hline
% {GPT}         &  56  & {60} & {96} \\
% {Mixtral}         & {60} & {61} & {78} \\
% {Gemini} & {63} & {61} & {96} \\
% \end{tabular}
% }
% \caption{Percentage of success of expected output languages for each model/task}
% \label{tab:output_error}
% \end{table}

\begin{figure*}[]
    \centering
    \subfloat[QA\label{fig:otuput_qa}]{{\includegraphics[width=0.23\textwidth]{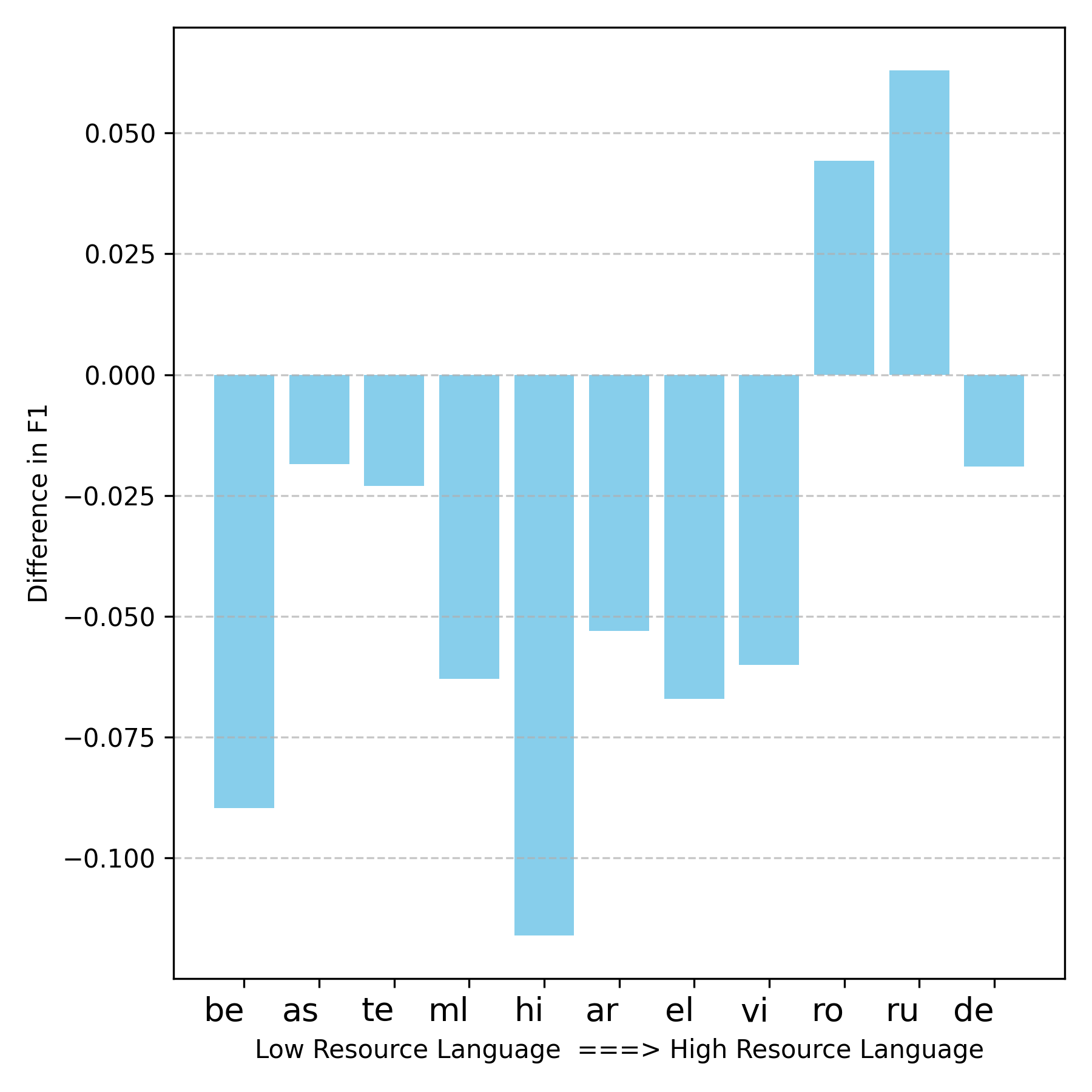} }}%
    \hfill
    \subfloat[NER\label{fig:family_langs_1}]{{\includegraphics[width=0.23\textwidth]{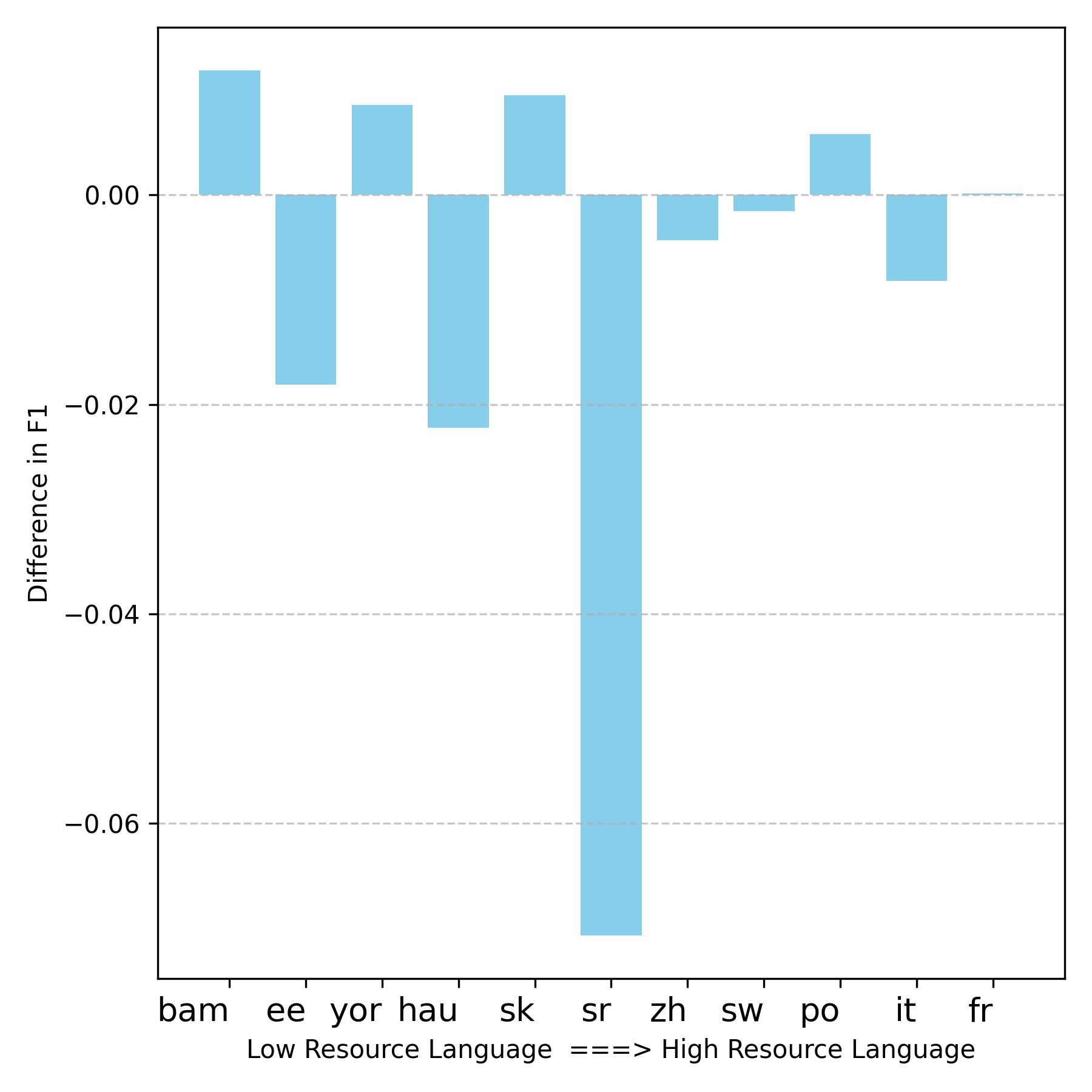} }}%
    \hfill
    \subfloat[Summarization\label{fig:family_langs_2}]{{\includegraphics[width=0.23\textwidth]{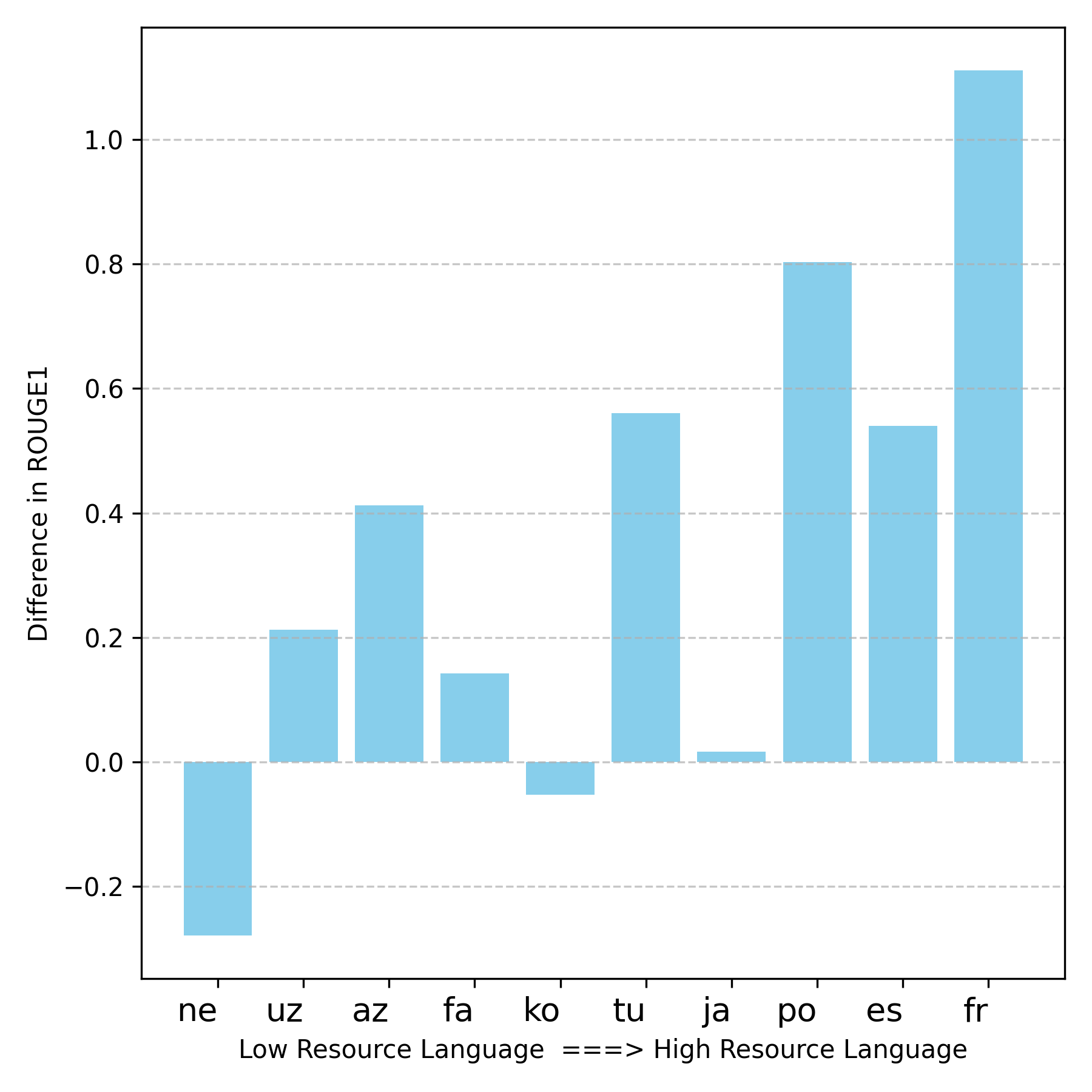} }}%
    \caption{Output Performance Gap (English - Source) for each task/language}%
    \label{fig:output_english_output_target}
\end{figure*}

\subsection{Prompting}

\label{appendix:prompts}

\paragraph{Question Answering}

Answer the following <Question> based only on the given <Context>. Follow these instructions:
\begin{itemize}
    \item Include only words from the given context in your answer.
    \item Keep the answer as short as possible.
    \item Provide the answer in \textit{expected output language}.
\end{itemize}

\paragraph{Named Entity Recognition}

You are an NLP assistant whose purpose is to perform Named Entity Recognition (NER). 
You need to assign each entity a tag from the following:
\begin{enumerate}
    \item PER means a person.
    \item ORG means an organization.
    \item LOC means a location entity.
\end{enumerate}
The output should be a list of tuples in the format:
\[
[(\text{Tag}, \text{Entity}), (\text{Tag}, \text{Entity})]
\]
for each entity in the sentence. The entities should be in the \textit{expected output language}.

\paragraph{Summarization}

Write a summary of the given <Text>
The output should be in \textit{expected output language}.
The output must be up to 2 sentences maximum.

\paragraph{Natural Language Inference}
You are an NLP assistant whose purpose is to solve Natural Language Inference (NLI) problems.
NLI is the task of determining the inference relation between two texts: entailment, contradiction, or neutral.
Your answer should be one word from the following: entailment, contradiction, or neutral.

% \paragraph{Improvement Over Pre-Translation Configuration}

% Figure \ref{fig:improvement_over_english} illustrates the improvement percentage over \emph{pre-translation}.
% The figure reveals that using \emph{Selective Pre-Translation} has a positive impact on more than 90\% of the selected languages. In addition, this approach has a greater positive impact on languages with smaller pre-training data. 
% % Consequently, both figures highlight that low-resource languages require a more complex configuration, which combines both the source and English languages. This superiority is reflected in extremely low-resource languages (red bars), which demonstrate greater improvement over other resource classes.

% \label{appendix:improvement_over_base_configurations}

\subsubsection{The Holy Grail of Optimal Configuration}
\label{appendix:optimal_configuration}

\paragraph{Few-Shot Examples Impact}

Figure \ref{fig:zero_shot_few_shot} demonstrates that for all tasks, using a few-shot setting over a zero-shot setting yields better results. Interestingly, For all tasks, except for NLI, high-resource languages achieved better improvement when considering a few-shot setting over low-resource languages.

\paragraph{Output Selection Effects}

Figure \ref{fig:output_english_output_target} demonstrates that while in extractive QA the output should be in the source language, and in the summarization task, the output should be in English; in NER, the output is ambiguous.

\label{appendix:output_english_output_target_appendix}

\begin{figure*}[]
    \centering
    \includegraphics[width=0.8\linewidth, height=0.11\textheight]{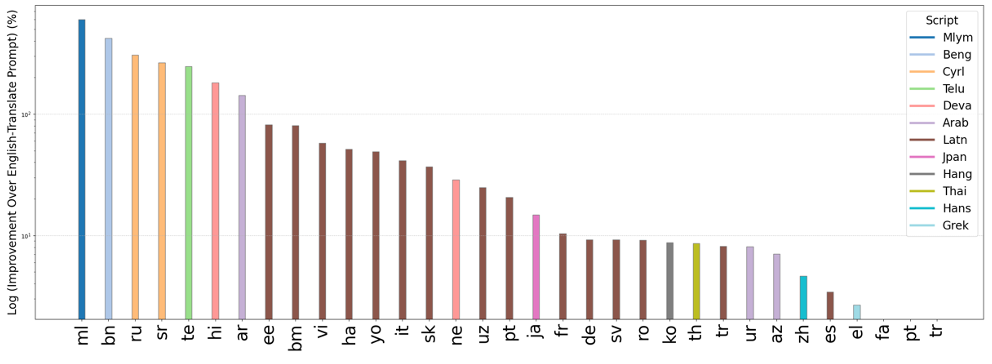}
    \caption{Percentage improvement over \emph{pre-translation} approach, when using the highest configuration for each task For GPT-3.5-Turbo. The bars are color-coded based on the language family script.}
    \label{fig:family_scripts}
\end{figure*}

\begin{CJK*}{UTF8}{gbsn}

\begin{table*}[h]
\centering
\scalebox{0.4}{
\begin{tabular}{lllll}
\multicolumn{1}{c}{\textit{\textbf{Task}}} & \multicolumn{1}{c}{\textit{\textbf{Model Output}}}                                        & \multicolumn{1}{c}{\textit{\textbf{Expected Output}}}  & \multicolumn{1}{c}{\textit{\textbf{Explnation}}}                                                               & \multicolumn{1}{c}{\textit{\textbf{Phenoenen}}} \\ \hline
\multirow{5}{*}{\textbf{NER}}              & {[} 'LOC: 新 北 市', 'LOC: 平 溪 區' {]}                                                        & {[}(LOC, '新 北 市',), (LOC, '平 溪 區',){]}                 & List of strings, instead of list of tuples.                                                                    & \textbf{Format Inconsistency}                   \\ \cline{2-5} 
                                           & {[}PER: Hiei{]}\textbackslash n- {[}PER: Hinata{]}\textbackslash n                       & {[}'PER', 'Hiei'), ('PER', 'Hinata'){]}                & New line between each entity.                                                              & \textbf{Format Inconsistency}                   \\ \cline{2-5} 
                                           & Ner Tags: {[}'PER: LL Cool J'{]}                                                          & {[}(PER: LL Cool J){]}                                 & Redundant words in the beginning.                                                                              & \textbf{Extraneous information}                \\ \cline{2-5} 
                                           & {[}{]} (No entities found in the sentence)                                                & {[}{]}                                                 & Redundant words in the end.                                                                                    & \textbf{Extraneous information}                \\ \cline{2-5} 
                                           & Since the last sentence is in English, I will provide the NER tags in English as well     & {[}(PER: Давид IV Грузијски){]}                          & Refusing to output in the desired language.                                                                                 & \textbf{Unwarranted Refusal}                    \\ \hline
\multirow{2}{*}{\textbf{QA}}               & {[}The united states{]}                                                                   & The united states                                      & List of string instead of a string.                                                                            & \textbf{Format Inconsistency}                   \\ \cline{2-5} 
                                           & The question cannot be answered as the answer is not provided in the given context        & {[}Luke Kuechly{]}                                     & Insufficent information                                                                     & \textbf{Unwarranted Refusal}                    \\ \hline
\multirow{2}{*}{\textbf{NLI}}              & The second statement neutral because it does not provide any information that contradicts & neutral                                                & Unnecessary justification for the choice.                                                                      & \textbf{Extraneous information}                \\ \cline{2-5} 
                                           & vinculación                                                                               & entailment                                             & Spanish word for entailment instead of English.                                                                & \textbf{Wrong Language}                         \\ \hline
\textbf{Summarization}                     & Resumo: O ministro de Emergências da Rússia, Sergei Shoigu ...                            & O ministro de Emergências da Rússia, Sergei Shoigu ... & \begin{tabular}[c]{@{}l@{}}Redundant words ('Resumo' - Summary in Portuguese)\\  in the beginning.\end{tabular} & \textbf{Extraneous information}                \\ \hline
\end{tabular}%
}
\caption{Error analysis of unexpected model outputs and observed in various tasks/languages.}
\label{tab:qualitative_analysis}

\end{table*}

\end{CJK*}

\begin{table}[]
\centering
\scalebox{0.9}{
\begin{tabular}{lcccl }
\Xhline{6\arrayrulewidth}
\textbf{Model}        & \textbf{QA} & \textbf{NER} & \textbf{Summarization} \\ \hline
{GPT}         &  56  & {60} & {96} \\
{Mixtral}         & {60} & {61} & {78} \\
{Gemini} & {63} & {61} & {96} \\
\end{tabular}
}
\caption{Percentage of success of expected output languages for each model/task}
\label{tab:output_error}
\end{table}

\subsubsection{Factors Explaining Performance}

\subsubsection{Script Impact}
\label{appendix:script_impact}
Figure \ref{fig:family_scripts} presents the performance improvement achieved by the highest-performing prompt configuration among all configurations compared to the pre-translation prompt, for each language. Notably, the language family (as categorized by scripts) reveals a relatively even distribution of performance gains within the same language family. For example, languages using the Cyrillic script show greater improvement than those using the Latin script. 
Interestingly, languages in the same script family sometimes show varying results; for example, Spanish and Ewe belong have Latin script, but Ewe shows greater improvement over Spanish.

\paragraph{Linguistic Similarity To English}
We used the lang2vec\footnote{\url{https://github.com/antonisa/lang2vec}} library to obtain syntactic similarity scores for each language. The Pearson correlation was calculated based on two vectors: one representing language similarities (ranging from 0 to 1) and the other representing model performance scores for each language across tasks. This correlation was calculated at the instance level. Table \ref{tab:syntactic_heatmaps} shows a positive correlation between model performance and syntactic similarity to English, especially for the summarization task, indicating that syntactic similarity to English significantly improves performance in this task. Additionally, NER also exhibits positive correlations, suggesting that models can better identify and classify entities in languages that share syntactic features with English.
\label{appendix:factors_explaning_performance}

\section{Error Analysis}
\label{appendix:error_analysis}

\subsection{Format Issues}

Automatic evaluation requires consistent output formatting, 
especially in tasks like Named Entity Recognition (NER), which must adhere to a predefined format rather than free text. A common practice involves prompting the model to generate results in a specific format, such as a list of tuples representing entities and their types (e.g., \((Loc, New York City)\). However, achieving perfect consistency can be challenging. Models may not always adhere to the requested format, leading to difficulties in evaluation.

\paragraph{Qualitative Analysis}

We analyzed unexpected model outputs in various tasks and languages. For each task, we noted common phenomena observed and the expected model output. The results in Table \ref{tab:qualitative_analysis} reveal that for the NER task, due to its rigid format, the model exhibited many error types. The models showed phenomena such as format inconsistency and extraneous introduction, which require a more generative normalization method to handle. An interesting phenomenon that made our modular selective pre-translating approach difficult to implement is unwarranted refusal, where the model refuses to output in the required language.

\subsection{Incorrect Output Language }

Table \ref{tab:output_error} summarizes the percentage of accurately outputted language for all tasks (except NLI, due to its index-based format) across all models. The results reveal that in extractive tasks such as extractive QA and NER, where the output overlaps with the context, the model struggles the most to output in the desired language. However, in abstractive summarization, a generation task, the model had better success.

\begin{table}[]
\centering
\scalebox{0.9}{
\begin{tabular}{lcccl }
\Xhline{6\arrayrulewidth}
\textbf{Model}        & \textbf{QA} & \textbf{NER} & \textbf{Summarization} & \textbf{NLI} \\ \hline
GPT                                                                                       & 0.14*      & 0.28**      & 0.42**                & 0.01        \\
Mixtral                                                                                   & 0.13*      & 0.2**       & 0.31**                & 0.08         \\
Gemini                                                                                    & 0.1*       & 0.19**      & 0.25**                & 0.01        
\end{tabular}%
}
\caption{Pearson correlation between linguistic syntactic
similarity to English and task performance for GPT,
Mixtral, and Gemini. * p < 0.05, ** p < 0.01}
\label{tab:syntactic_heatmaps}
\end{table}

\section{Translation: Key to Pre-Translation}

\subsection{Machine Translation Engines Comparison}

% To assess which machine translation tool to use we used Google Translate API and Bing Translate. We didn't consider multilingual LLMs for machine translation  because we are based on the finding of \emph{zhu2023multilingual} which found that multilingual LLMS for MT still face a large gap towards the commercial translation system like Google Translate, especially on low-resource languages. The results in Table \ref{tab:google_vs_bing} and Figure \ref{fig:google_bing} demonstrate that Google Translate API outperformed Bing Translator in all the evaluated metrics, highlighting its high performance. Interestingly, the languages that achieved the highest scores are Welsh and Maltese which are both considered low-resource languages. 
% \label{appendix:mt_engines_comparsion}

\begin{figure*}[t]
    \centering
    \subfloat[BLEU\label{fig:otuput_qa}]{{\includegraphics[width=1\textwidth]{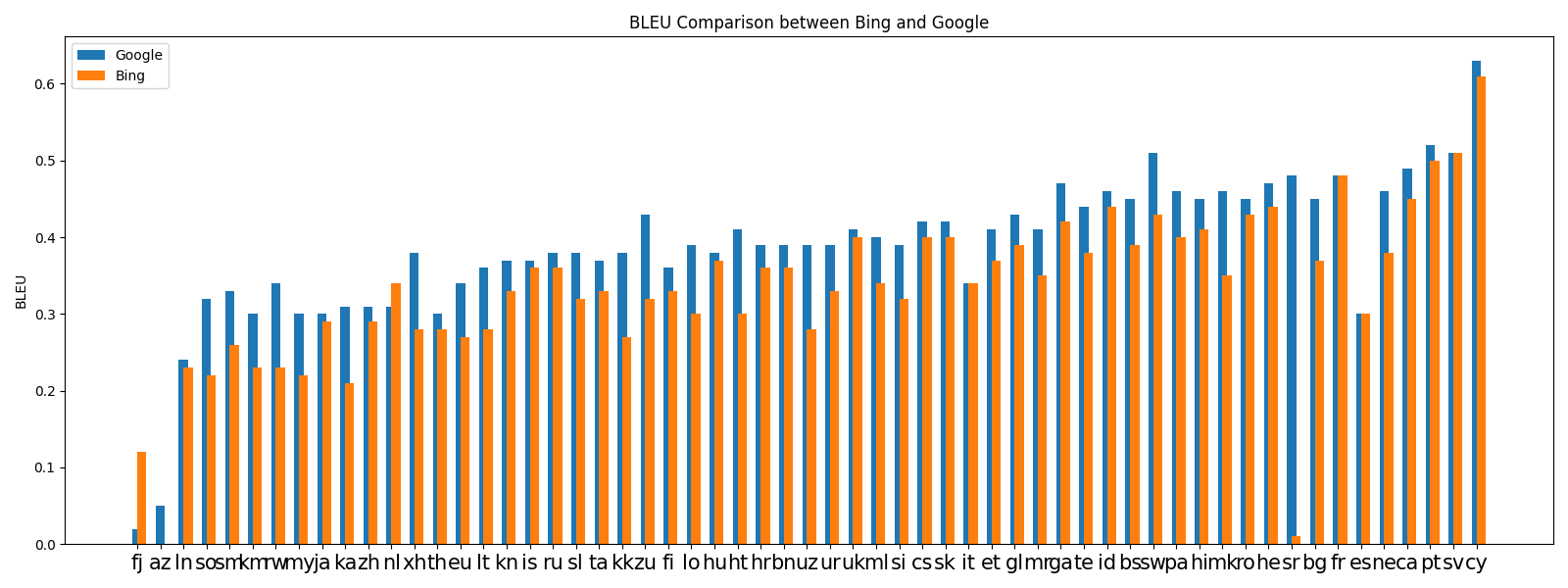} }}%
    \hfill
    \subfloat[ROUGE\label{fig:family_langs_1}]{{\includegraphics[width=1\textwidth]{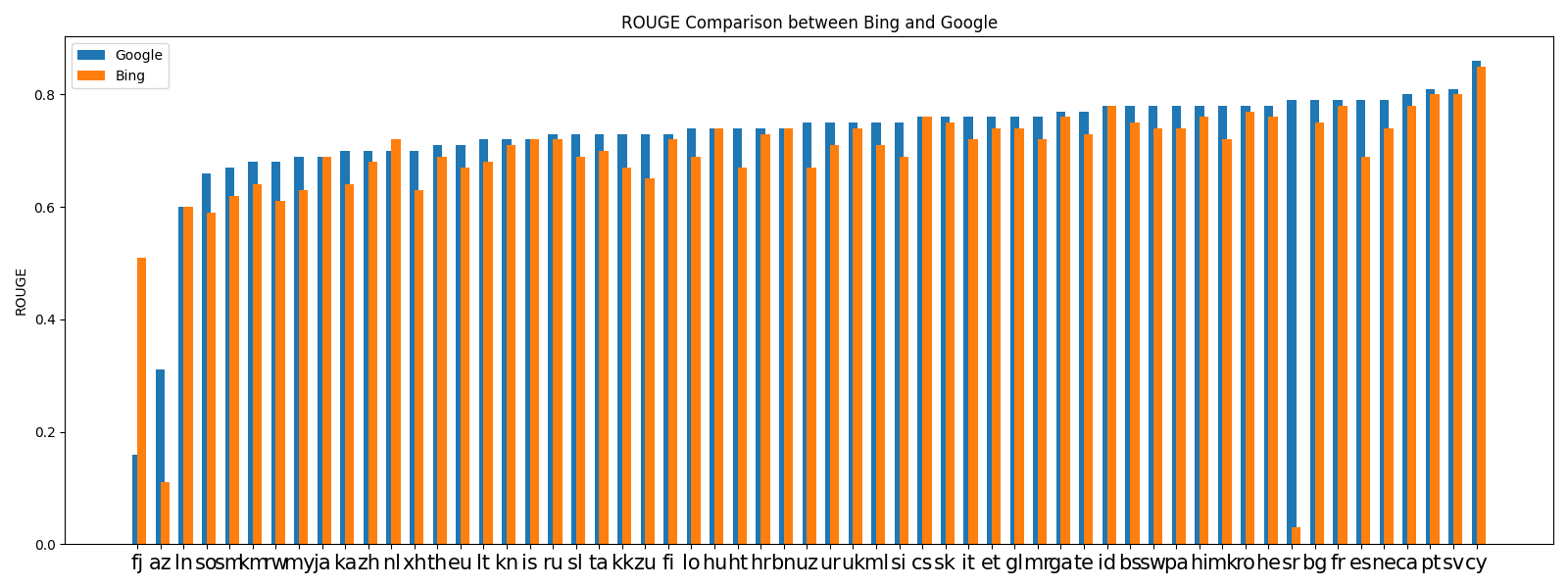} }}%
    \hfill
    \subfloat[Meteor\label{fig:family_langs_2}]{{\includegraphics[width=1\textwidth]{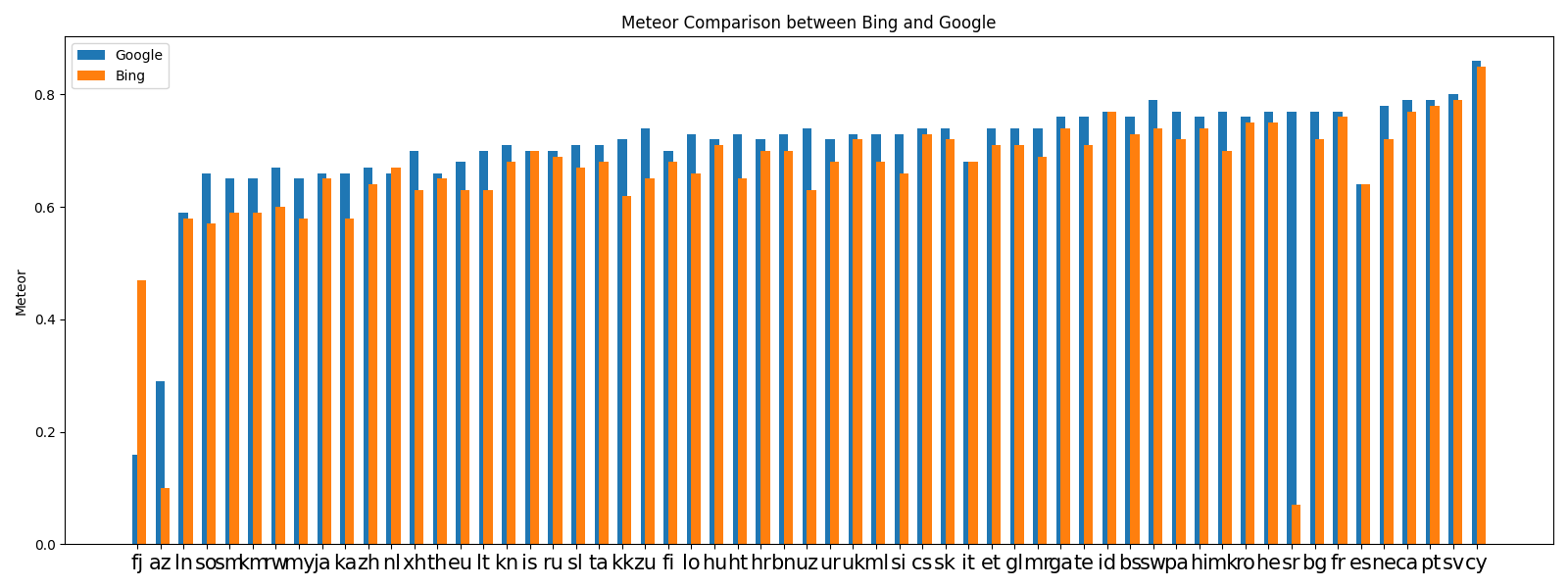} }}%
    \caption{Google Translate API vs Bing Translator Comparsion}%
    \label{fig:google_bing}
\end{figure*}

To evaluate machine translation tools, we compared Google Translate API and Bing Translator. We excluded multilingual LLMs from consideration, as \emph{zhu2023multilingual} found that these models still lag behind commercial systems like Google Translate, especially for low-resource languages. As shown in Table \ref{tab:google_vs_bing} and Figure \ref{fig:google_bing}, Google Translate outperformed Bing Translator across all metrics, demonstrating superior performance. Notably, Welsh and Maltese, both low-resource languages, achieved the highest scores. \label{appendix:mt_engines_comparsion}.

\subsection{Linguistic Similarity To English}

The results in Table \ref{tab:correlation_subsets} demonstrate the correlation between the syntactic similarity to English of the language and the ROUGE translation score of the language. The results show that the most significant correlation was observed in languages belonging to the high-resource category, and this correlation decreases as the class of the language becomes low-resource.

\label{appendix:linguistic_similarity_to_english_correlation}

.
\section{Selective Pre-Translation Prompt Generator}
\label{appendix:hf_space}

\begin{figure*}[th]
    \centering
    \scalebox{1}{
    \includegraphics[width=\textwidth]{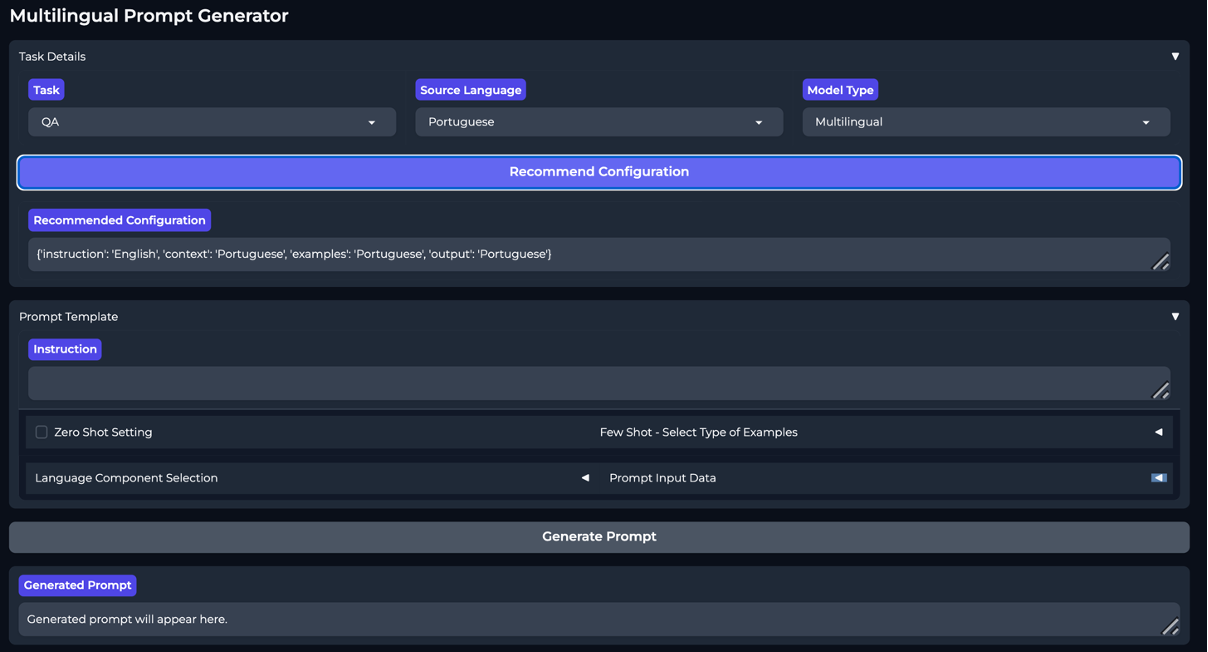}
    }
   \caption{\emph{Recomended Configuration:} Overview of our application. From top to bottom: Task Details—general configuration about the task (Task, Language, Model). Clicking on "Recommended Configuration" provides a suggested selective pre-translation configuration.}

    % Bar colors correspond to the log-normalized number of tokens in GPT-3 pre-training data.
    
    \label{fig:app}
\end{figure*}

\begin{figure*}[th]
    \centering
    \scalebox{1}{
    \includegraphics[width=\textwidth]{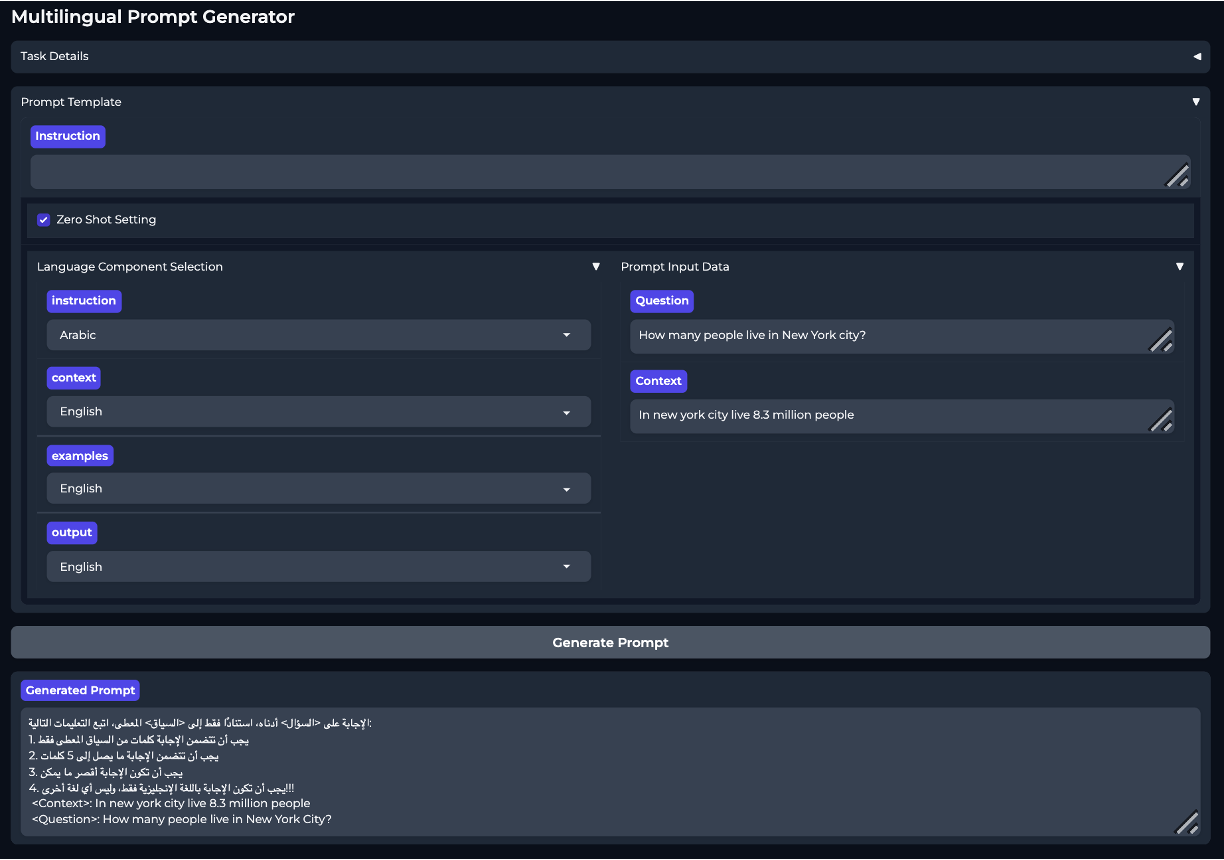}
    }
   \caption{\emph{Generating selective pre-translation prompt for zero-shot:}  The user needs to configure the instruction (optional) and the languages for the components under "Language Component Selection": instruction, context, examples, and output. Additionally, under "Prompt Input Data," the user must configure the relevant input data or task, such as the question and context for QA in this example. Clicking on "Generate Prompt" provides a zero-shot pre-translation prompt}

    % Bar colors correspond to the log-normalized number of tokens in GPT-3 pre-training data.
    
    \label{fig:zero_shot_app}
\end{figure*}

\begin{figure*}[th]
    \centering
    \scalebox{1}{
    \includegraphics[width=\textwidth]{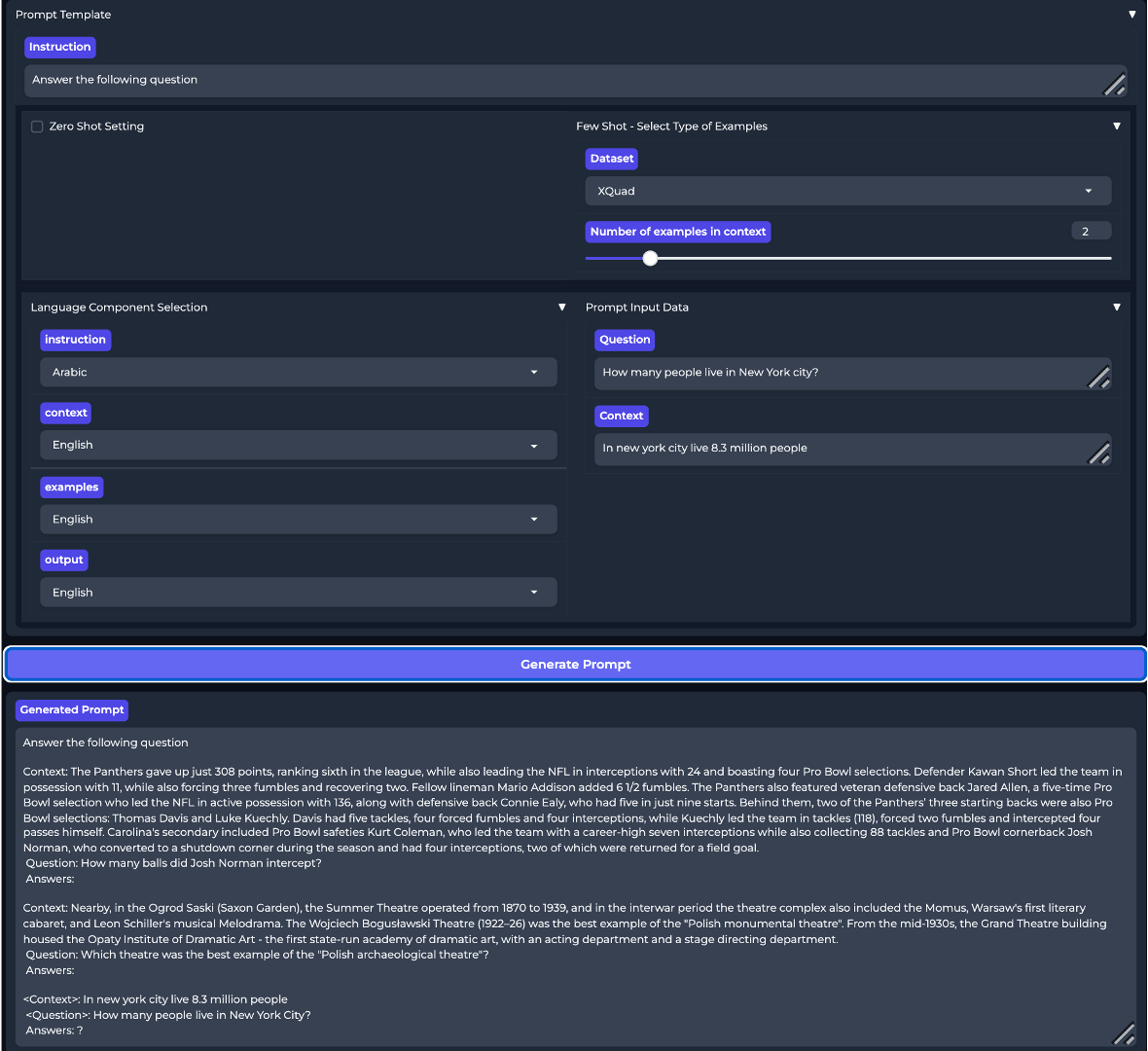}
    }
   \caption{\emph{Generating selective pre-translation prompt for few-shot:} Here, the user must also configure the few-shot settings: the dataset to use (from which the few-shot examples are taken) and the number of examples to use (default = 1).}
    \label{fig:few_shot_app}
\end{figure*}

We have launched a space on Hugging Face. The space makes it easy for the community to receive recommended configurations based on the type of task and language. In Figure \ref{fig:app}
, we can see an overview of the application and an example of a recommended configuration. Figures \ref{fig:zero_shot_app} and \ref{fig:few_shot_app} provide examples of generating prompts for zero-shot and few-shot settings.

\section{Detailed Results}
\label{appendix:detailed_results}
The results across all tasks, languages and models are included in our benchmarking exercise are
provided in Table \ref{tab:qa_quad} (for XQuAD), \ref{tab:qa_indicqa} (for indciQA), \ref{tab:wikiann} (for WikiANN), \ref{tab:masakhaner} (for MasakhNER), \ref{tab:xlsum} (for XL-Sum), \ref{tab:xnli} (for XNLI).
The result of the correlation for Gemini are included in Table \ref{tab:gemini_correlation}, for Mixtral in Table \ref{tab:mixtral_correlation}, and for bloomz in Table \ref{tab:bloomz_correlation}
\label{appendix:detailed_results}

\begin{table}[]
\centering
\scalebox{0.7}{
\begin{tabular}{lcccl }
\Xhline{3\arrayrulewidth}
\textbf{Group}        & \textbf{Pearson Correlation} & \textbf{P-value} \\ \hline
{High Resource}         &  0.73  & {0.05}                      \\
{Medium Resource}         & {0.48}                          & {0.07}                      \\
{Low Resource} & {0.06}                           & {0.78}                      \\
{Extremly Low Resource} & {-0.34}                           & {0.30}                      \\

\end{tabular}
}
\caption{Correlation between syntactic similarity to English and the ROUGE score (by language subset).}
\label{tab:correlation_subsets}

\end{table}

\begin{table}[]
\centering
\resizebox{\columnwidth}{!}{%
\begin{tabular}{llll|lll}
                        & \multicolumn{3}{c|}{\textbf{\underline{Google Translate API}}} & \multicolumn{3}{c}{\textbf{\underline{Bing Translator}}}       \\
Language                & \textit{\textbf{ROUGE}} & \textit{\textbf{Meteor}} & \textit{\textbf{BLEU}} & \textit{\textbf{ROUGE}} & \textit{\textbf{Meteor}} & \textit{\textbf{BLEU}} \\
Welsh                   & 0.86                    & 0.86                     & 0.63                   & 0.85                    & 0.85                     & 0.61                   \\
Maltese                 & 0.84                    & 0.83                     & 0.59                   & 0.83                    & 0.82                     & 0.56                   \\
Danish                  & 0.81                    & 0.79                     & 0.51                   & 0.81                    & 0.79                     & 0.49                   \\
Swedish                 & 0.81                    & 0.80                     & 0.51                   & 0.80                    & 0.79                     & 0.51                   \\
Portuguese              & 0.81                    & 0.79                     & 0.52                   & 0.80                    & 0.78                     & 0.50                   \\
Catalan                 & 0.80                    & 0.79                     & 0.49                   & 0.78                    & 0.77                     & 0.45                   \\
\hline
\end{tabular}%
}
\caption{Comparsion between Google Translate API and Bing Translator.}
\label{tab:google_vs_bing}
\end{table}

\begin{table}[!th]
\resizebox{\columnwidth}{!}{%
\begin{tabular}{@{}ccccc@{}}
\toprule
\textbf{Language}    & \textbf{Lang Code} & \textbf{Number of Tokens (M)} & \textbf{Percentage of Tokens} & \textbf{Class}\\ \midrule
English     & en & 181,015 & 92.64\%  & A             \\ \midrule
French      & fr &  \pz3,553   &  1.81853\% & A             \\
German      & de &  \pz2,871   &  1.46937\%  & A            \\
Spanish     & es &  \pz1,510   &  0.77289\% & A             \\
Italian     & it &  \pz1,188   &  0.60793\% & A             \\
Portuguese  & po &  \pz1,025   &  0.52483\% & A             \\
Russian     & ru &  \pzz368   &  0.18843\% & A             \\
\hline
\end{tabular}%
}
\caption{ List of languages, language codes, number of tokens in pre-trained GPT-3 data, data ratios. The languages are grouped into four classes based on their data
ratios in the GPT-3 pre-trained data: High Resource
(H > 0.1\%), Medium Resource (M > 0.01\%), and Low
Resource (L < 0.01\%), and extremely low resource for unrepresented languages.}
\label{tab:languages_classes}
\end{table}

\begin{table*}[] 
\centering
\resizebox{\textwidth}{!} {
\begin{tabular}{lllllllllllllllllll}
\multicolumn{4}{c}{\textbf{Configuration}}                                                    & \multicolumn{3}{c}{\textbf{Assamese}}             & \multicolumn{3}{c}{\textbf{Bengali}}              & \multicolumn{3}{c}{\textbf{Hindi}}                & \multicolumn{3}{c}{\textbf{Malayalam}}            & \multicolumn{3}{c}{\textbf{Telugu}}               \\
\multicolumn{1}{c}{P} & \multicolumn{1}{c}{I} & \multicolumn{1}{c}{C} & \multicolumn{1}{c}{O} & \textit{gemini} & \textit{gpt} & \textit{mixtral} & \textit{gemini} & \textit{gpt} & \textit{mixtral} & \textit{gemini} & \textit{gpt} & \textit{mixtral} & \textit{gemini} & \textit{gpt} & \textit{mixtral} & \textit{gemini} & \textit{gpt} & \textit{mixtral} \\ \hline
S                     & S                     & Z                     & E                     & 0.2             & 0.43         & 0.00             & 0.16            & 0.48         & 0.02             & 0.26            & 0.22         & 0.19             & 0.19            & 0.25         & 0.03             & 0.29            & 0.15         & 0.13             \\
E                     & S                     & Z                     & E                     & 0.15            & 0.03         & 0.00             & 0.11            & 0.03         & 0.02             & 0.21            & 0.30         & 0.12             & 0.13            & 0.13         & 0.14             & 0.12            & 0.10         & 0.10             \\
S                     & S                     & S                     & E                     & 0.59            & 0.35         & 0.10             & 0.65            & 0.52         & 0.28             & 0.74            & 0.68         & 0.37             & 0.46            & 0.36         & 0.04             & 0.53            & 0.23         & 0.19             \\
\hline
\end{tabular}%
}
\caption{Comparing performance of different models on all languages in IndicQA. Metric: F1 Score.}
\label{tab:qa_indicqa}
\end{table*}

\begin{table*}[!h] 
\centering
\resizebox{\textwidth}{!} {
\begin{tabular}{llllllllllllllllllllllllllllllll}
\multicolumn{3}{c}{\textbf{Configuration}}                                                       & \multicolumn{3}{c}{\textbf{Arabic}} & \multicolumn{3}{c}{\textbf{Chinese}} & \multicolumn{2}{c}{\textbf{Greek}} & \multicolumn{3}{c}{\textbf{Hindi}} & \multicolumn{3}{c}{\textbf{Spanish}} & \multicolumn{3}{c}{\textbf{Swahili}} & \multicolumn{3}{c}{\textbf{Thai}} & \multicolumn{3}{c}{\textbf{Turkish}} & \multicolumn{3}{c}{\textbf{Urdu}} & \multicolumn{3}{c}{Bulgarian} \\
\multicolumn{1}{c}{\textit{P}} & \multicolumn{1}{c}{\textit{I}} & \multicolumn{1}{c}{\textit{C}} & gemini     & gpt      & mixtral     & gemini      & gpt      & mixtral     & gemini            & gpt            & mixtral     & gemini     & gpt     & gemini      & gpt      & mixtral     & gemini      & gpt      & mixtral     & gemini     & gpt     & mixtral    & gemini      & gpt      & mixtral     & gemini     & gpt     & mixtral    & gemini   & gpt    & mixtral   \\
E                              & S                              & E                              & 0.72       & 0.68     & 0.51        & 0.63        & 0.58     & 0.55        & 0.75              & 0.53           & 0.61        & 0.59       & 0.49    & 0.59        & 0.56     & 0.32        & 0.65        & 0.54     & 0.54        & 0.52       & 0.71    & 0.33       & 0.45        & 0.45     & 0.0         & 0.66       & 0.52    & 0.41       & 0.49     & 0.54   & 0.46      \\
S                              & S                              & E                              & 0.65       & 0.67     & 0.54        & 0.62        & 0.63     & 0.52        & 0.66              & 0.59           & 0.54        & 1.00       & 0.46    & 0.6         & 0.64     & 0.52        & 0.62        & 0.64     & 0.59        & 0.52       & 0.56    & 1.0        & 0.47        & 0.42     & 0.5         & N.A        & 0.52    & N.A        & 0.52     & 0.54   & 0.44      \\
\end{tabular}%
}
\caption{Comparing performance of different models on all languages in XNLI. Metric: Acc Score.}
\label{tab:xnli}
\end{table*}

\begin{table*}[] 
\centering
\resizebox{\textwidth}{!} {
\begin{tabular}{llllllllllllllllllllll}
\multicolumn{4}{c}{\textbf{Configuration}}                                                                                        & \multicolumn{3}{c}{\textbf{Arabic}}                                                                           & \multicolumn{3}{c}{\textbf{German}}                                                                           & \multicolumn{3}{c}{\textbf{Greek}}                                                                            & \multicolumn{3}{c}{\textbf{Romanian}}                                                                         & \multicolumn{3}{c}{\textbf{Russian}}                                                                          & \multicolumn{3}{c}{\textbf{Vietnamese}}                                                                       \\
\multicolumn{1}{c}{\textit{P}} & \multicolumn{1}{c}{\textit{I}} & \multicolumn{1}{c}{\textit{C}} & \multicolumn{1}{c}{\textit{O}} & \multicolumn{1}{c}{\textit{gemini}} & \multicolumn{1}{c}{\textit{gpt}} & \multicolumn{1}{c}{\textit{mixtral}} & \multicolumn{1}{c}{\textit{gemini}} & \multicolumn{1}{c}{\textit{gpt}} & \multicolumn{1}{c}{\textit{mixtral}} & \multicolumn{1}{c}{\textit{gemini}} & \multicolumn{1}{c}{\textit{gpt}} & \multicolumn{1}{c}{\textit{mixtral}} & \multicolumn{1}{c}{\textit{gemini}} & \multicolumn{1}{c}{\textit{gpt}} & \multicolumn{1}{c}{\textit{mixtral}} & \multicolumn{1}{c}{\textit{gemini}} & \multicolumn{1}{c}{\textit{gpt}} & \multicolumn{1}{c}{\textit{mixtral}} & \multicolumn{1}{c}{\textit{gemini}} & \multicolumn{1}{c}{\textit{gpt}} & \multicolumn{1}{c}{\textit{mixtral}} \\
\hline
\end{tabular}%
}

\captionsetup{justification=centering} % Centering the caption
\caption{Comparing performance of different models on all languages in XQuAD. Metric: F1 Score.}
\label{tab:qa_quad}
\end{table*}

\begin{table*}[!ht] 
\centering
\resizebox{\textwidth}{!} {
\begin{tabular}{lllllllllllllllllllllllll}
\multicolumn{4}{c}{\textbf{Configuration}}                                                                                        & \multicolumn{3}{c}{\textbf{Chinese}}              & \multicolumn{3}{c}{French}                        & \multicolumn{3}{c}{\textbf{Italian}}              & \multicolumn{3}{c}{\textbf{Portuguese}}           & \multicolumn{3}{c}{\textbf{Serbian}}              & \multicolumn{3}{c}{\textbf{Slovak}}               & \multicolumn{3}{c}{Swedish}                       \\
\multicolumn{1}{c}{\textit{P}} & \multicolumn{1}{c}{\textit{I}} & \multicolumn{1}{c}{\textit{C}} & \multicolumn{1}{c}{\textit{O}} & \textit{gemini} & \textit{gpt} & \textit{mixtral} & \textit{gemini} & \textit{gpt} & \textit{mixtral} & \textit{gemini} & \textit{gpt} & \textit{mixtral} & \textit{gemini} & \textit{gpt} & \textit{mixtral} & \textit{gemini} & \textit{gpt} & \textit{mixtral} & \textit{gemini} & \textit{gpt} & \textit{mixtral} & \textit{gemini} & \textit{gpt} & \textit{mixtral} \\ \hline
E                              & E                              & E                              & E                              & 0.00            & 0.01         & 0.23             & 0.53            & 0.58         & 0.15             & 0.38            & 0.53         & 0.23             & 0.56            & 0.60         & 0.41             & 0.20            & 0.21         & 0.00             & 0.48            & 0.53         & 0.19             & 0.59            & 0.62         & 0.41             \\
\hline
\end{tabular}%
}
\caption{Comparing performance of different models on all languages in WikiANN. Metric: F1 Score.}
\label{tab:wikiann}
\end{table*}

\begin{table*}[!h] 
\centering
\resizebox{\textwidth}{!} {
\begin{tabular}{llllllllllllllll}
\multicolumn{4}{c}{\textbf{Configuration}}                                                    & \multicolumn{3}{c}{\textbf{Bambara}}              & \multicolumn{3}{c}{\textbf{Ewe}}                  & \multicolumn{3}{c}{\textbf{Hausa}}                & \multicolumn{3}{c}{\textbf{Yoruba}}               \\
\multicolumn{1}{c}{P} & \multicolumn{1}{c}{I} & \multicolumn{1}{c}{C} & \multicolumn{1}{c}{O} & \textit{gemini} & \textit{gpt} & \textit{mixtral} & \textit{gemini} & \textit{gpt} & \textit{mixtral} & \textit{gemini} & \textit{gpt} & \textit{mixtral} & \textit{gemini} & \textit{gpt} & \textit{mixtral} \\ \hline
E                     & E                     & E                     & E                     & 0.16            & 0.18         & 0.10             & 0.42            & 0.47         & 0.22             & 0.45            & 0.24         & 0.26             & 0.06            & 0.09         & 0.03             \\
E                     & E                     & E                     & S                     & 0.17            & 0.17         & 0.07             & 0.43            & 0.48         & 0.23             & 0.46            & 0.61         & 0.31             & 0.07            & 0.09         & 0.05             \\
E                     & E                     & S                     & E                     & 0.15            & 0.17         & 0.12             & 0.44            & 0.52         & 0.22             & 0.45            & 0.52         & 0.20             & 0.07            & 0.08         & 0.04             \\
E                     & E                     & S                     & S                     & 0.14            & 0.15         & 0.07             & 0.39            & 0.5          & 0.17             & 0.43            & 0.39         & 0.10             & 0.06            & 0.08         & 0.03             \\
\hline
\end{tabular}%
}
\caption{Comparing performance of different models on all languages in MasakhaNER. Metric: F1 Score.}
\label{tab:masakhaner}
\end{table*}

\begin{table*}[!h] 
\centering
\resizebox{\textwidth}{!} {
\begin{tabular}{llllllllllllllllllllllllllllllllll}
\multicolumn{4}{c}{\textbf{Configuration}}                                                                                        & \multicolumn{3}{l}{\textbf{Azerbaijani}}           & \multicolumn{3}{l}{\textbf{French}}               & \multicolumn{3}{l}{\textbf{Japanese}}             & \multicolumn{3}{l}{\textbf{Korean}}               & \multicolumn{3}{l}{\textbf{Nepali}}                            & \multicolumn{3}{l}{\textbf{Persian}}                     & \multicolumn{3}{l}{\textbf{Portuguese}}           & \multicolumn{3}{l}{\textbf{Spanish}}                     & \multicolumn{3}{l}{\textbf{Turkish}}                           & \multicolumn{3}{l}{\textbf{Uzbek}}                                 \\
\multicolumn{1}{c}{\textit{P}} & \multicolumn{1}{c}{\textit{I}} & \multicolumn{1}{c}{\textit{C}} & \multicolumn{1}{c}{\textit{O}} & \textit{gpt} & \textit{mixtral} & \textit{mixtral} & \textit{gemini} & \textit{gpt} & \textit{mixtral} & \textit{gemini} & \textit{gpt} & \textit{mixtral} & \textit{gemini} & \textit{gpt} & \textit{mixtral} & \textit{gemini}        & \textit{gpt} & \textit{mixtral}       & \textit{gemini}        & \textit{gpt} & \textit{mixtral} & \textit{gemini} & \textit{gpt} & \textit{mixtral} & \textit{gemini}        & \textit{gpt} & \textit{mixtral} & \textit{gemini}        & \textit{gpt} & \textit{mixtral}       & \textit{gemini}        & \textit{gpt}           & \textit{mixtral} \\
\hline
\end{tabular}%
}
\caption{Comparing performance of different models on all languages in XlSUM. Metric: ROUGE1 Score.}
\label{tab:xlsum}
\end{table*}

\begin{table*}[t]
\centering

\resizebox{\textwidth}{!}{%
\begin{tabular}{lllllllllllllllllll}
\toprule
\multicolumn{5}{c}{\textbf{Question Answering}}                   & \multicolumn{5}{c}{\textbf{Summarization}}                      & \multicolumn{5}{c}{\textbf{Named Entity Recognition}}            & \multicolumn{3}{c}{\textbf{NLI}} &          \\
code & instruction & context           & Examples & output          & code & instruction & context & Examples        & output           & code & instruction & context           & Examples         & output &       & instruction   & context    & Examples \\
ar   & -0.02       & \textbf{0.30**} & -0.10**  & \textbf{0.24**} & az   & -0.14**     & -0.01 & 0.04            & \textbf{-0.34**} & zh   & -0.07**     & \textbf{0.44**} & \textbf{-0.26**} & 0.00   & ar    & -0.02         & -0.01(   & -0.05    \\
as   & -0.05       & \textbf{0.35**} & -0.00    & \textbf{0.30**} & fr   & -0.07*      & 0.03  & \textbf{0.19**} & -0.04*           & fr   & 0.01        & 0.10            & -0.11*           & -0.01  & bu    & -0.00         & -0.02    & 0.03     \\
be   & -0.10*      & \textbf{0.39**} & -0.00    & \textbf{0.30**} & ja   & 0.15**      & -0.02 & \textbf{0.34**} & -0.07            & it   & 0.01        & 0.04            & 0.02             & -0.04  & zh    & 0.01          & -0.01    & -0.01    \\
ge   & 0.03        & 0.07*           & -0.09**  & 0.06            & ko   & 0.04        & -0.01 & \textbf{0.25**} & -0.06            & po   & 0.01        & 0.09*           & -0.15**          & 0.02   & ge    & 0.01          & 0.05     & -0.07    \\
gr   & -0.02       & \textbf{0.19**} & -0.09**  & 0.12**          & ne   & 0.00        & 0.02  & 0.13*           & -0.25**          & sr   & 0.05        & \textbf{0.44**} & \textbf{-0.26**} & 0.09   & gr    & 0.02          & -0.02    & 0.04     \\
\hline
\end{tabular}%
}
\caption{Point-biserial correlation of Gemini for each Language (denoted by ISO 639 code) nd each of the 4 prompt components - Instruction, Context, Examples, and Output. The p-value is given in the parentheses}
\label{tab:gemini_correlation}
\end{table*}

\begin{table*}[t]
\centering

\resizebox{\textwidth}{!}{%
\begin{tabular}{lllllllllllllllllll}
\toprule
\multicolumn{5}{c}{\textbf{Question Answering}}              & \multicolumn{5}{c}{\textbf{Summarization}}          & \multicolumn{5}{c}{\textbf{Named Entity Recognition}}             & \multicolumn{3}{c}{\textbf{NLI}} &         \\
code & instruction      & context           & examples & output & code & instruction    & context   & examples & output  & code & instruction & context            & examples          & output & code   & instruction   & context   & examples \\
ar   & 0.04             & 0.07            & -0.07** & 0.09** & az   & -0.09*         & -0.09   & 0.03    & -0.10   & zh   & 0.16        & \textbf{0.34**}  & \textbf{-0.29**} & -0.02  & ar     & -0.05         & 0.01    & 0.04    \\
as   & -0.16            & 0.02            & -0.01** & 0.04   & fr   & -0.07*         & 0.00    & 0.05    & -0.02   & fr   & 0.04        & 0.09             & 0.15**           & -0.04  & bu     & 0.01          & 0.01    & -0.03   \\
be   & -0.07            & \textbf{0.31**} & -0.07*  & 0.15** & ja   & -0.03          & 0.04    & 0.13*   & -0.18   & it   & -0.00       & 0.04             & 0.14**           & -0.01  & zh     & -0.03         & 0.00    & -0.01   \\
ge   & 0.04             & 0.07            & 0.13**  & -0.05  & ko   & 0.02           & -0.04   & 0.08    & -0.04   & po   & -0.01       & 0.08             & -0.02            & -0.01  & ge     & 0.02          & 0.02    & -0.01   \\
gr   & -0.06**          & 0.17            & 0.03    & 0.03   & ne   & -0.16          & 0.02    & 0.10    & -0.10   & sr   & 0.02        & \textbf{0.36**}  & 0.09**           & 0.08   & gr     & -0.06         & 0.01(   & -0.04   \\
hi   & -0.07**          & 0.17            & -0.05   & 0.10   & fa   & -0.09          & -0.00   & 0.14    & 0.05    & sk   & 0.05        & 0.11             & 0.12             & 0.06   & hi     & -0.01         & -0.01   & 0.01    \\
\hline
\end{tabular}%
}
\caption{Point-biserial correlation of Mixtral for each Language (denoted by ISO 639 code) nd each of the 4 prompt components - Instruction, context, Examples, and Output. The p-value is given in the parentheses}
\label{tab:mixtral_correlation}
\end{table*}

\begin{table*}[t]
\centering

\resizebox{\textwidth}{!}{%
\begin{tabular}{lllllllllllllllllll}
\hline
\multicolumn{5}{c}{{\textbf{Question Answering}}}     & \multicolumn{5}{c}{{\textbf{Summarization}}}               & \multicolumn{5}{c}{{\textbf{NER}}}                   & \multicolumn{4}{c}{{\textbf{NLI}}} \\
code & instruction & context          & Examples & output & code & instruction      & context          & Examples & output & code & instruction & context & Examples         & output & code  & instruction & context & output \\
ar   & 0.01        & \textbf{0.43**}  & N.A      & 0.02   & fr   & -0.06*           & 0.12**           & N.A      & -0.02  & zh   & 0.03        & 0.14    & \textbf{-0.16**} & 0.00   & ar    & 0.04**      & 0.05    & -0.07* \\
as   & 0.05        & \textbf{0.47**}  & N.A      & -0.03  & ja   & \textbf{-0.17**} & \textbf{-0.32**} & N.A      & 0.05   & fr   & -0.01       & 0.06    & \textbf{-0.35**} & -0.01  & zh    & 0.02        & -0.02   & -0.02  \\
be   & 0.001*      & \textbf{0.51**}  & N.A      & 0.04   & ne   & -0.05            & 0.04             & N.A      & 0.01   & it   & 0.04        & 0.02    & \textbf{-0.28**} & 0.01   & ge    & 0.06        & 0.04    & 0.02   \\
ge   & 0.05*       & -0.11**          & N.A      & 0.02   & po   & -0.01            & 0.04             & N.A      & -0.02  & po   & 0.02        & 0.04    & \textbf{-0.31**} & 0.01   & hi    & -0.06*      & 0.01    & -0.07  \\
gr   & 0.05*       & \textbf{-0.24**} & N.A      & ,0.11  & es   & 0.03             & -0.06            & N.A      & 0.01   & sr   & 0.05        & 0.01    & \textbf{-0.20**} & 0.021  & es    & 0.02        & -0.01*  & -0.01  \\
hi   & ,-0.00      & \textbf{0.44}    & N.A      & 0.01   & pe   & \textbf{-0.24*}  & \textbf{-0.25**} & N.A      & 0.06   & sk   & 0.01        & 0.02    & \textbf{-0.30**} & -0.02  & sw    & 0.07        & -0.01   & -0.03  \\
\hline
\end{tabular}
}
\caption{Point-biserial correlation of Bloomz for each Language (denoted by ISO 639 code) nd each of the 4 prompt components - Instruction, context, Examples, and Output. The p-value is given in the parentheses}
\label{tab:bloomz_correlation}
\end{table*}

\end{document}